\pdfoutput=1
\documentclass[11pt]{article}

\usepackage[final]{acl}

\usepackage{times}
\usepackage{latexsym}
\usepackage{subcaption}
\usepackage{tcolorbox}
\usepackage{booktabs}              % for professional-looking tables
\usepackage{amsmath}
\usepackage{fontawesome5} % For icons
\usepackage{tikz}
\usetikzlibrary{shapes.geometric, positioning, arrows.meta}
\usepackage{multirow}
\usepackage{amsfonts}
\usepackage{boldline}
\usepackage{array}
\usepackage{adjustbox}
\usepackage{booktabs}
\usepackage{makecell}
\usepackage{array}

\newcolumntype{C}[1]{>{\centering\arraybackslash}p{#1}}

\usepackage[T1]{fontenc}
\usepackage[utf8]{inputenc}

\usepackage{microtype}

\usepackage{inconsolata}

\usepackage{graphicx}

\definecolor{keywordcolor}{RGB}{0,0,139}
\definecolor{variablecolor}{RGB}{0,100,0}
\definecolor{green}{RGB}{0,255,0}
\definecolor{lightgreen}{RGB}{179, 247, 96}
\definecolor{blue}{RGB}{0,0,255}
\definecolor{orange}{RGB}{255,165,0}
\definecolor{red}{RGB}{255,0,0}
\definecolor{purple}{RGB}{128,0,128}
\definecolor{cyan}{RGB}{0,255,255}
\definecolor{magenta}{RGB}{255,0,255}
\definecolor{yellow}{RGB}{255,255,0}
\definecolor{brown}{RGB}{139,69,19}
\definecolor{gray}{RGB}{128,128,128}
\definecolor{pink}{RGB}{255,182,193}
\definecolor{teal}{RGB}{0,128,128}
\definecolor{olive}{RGB}{128,128,0}
\definecolor{lightblue}{RGB}{173,216,230}
\definecolor{darkblue}{RGB}{0,0,139}
\definecolor{customblue}{RGB}{149, 237, 231}

\definecolor{lightgreen}{rgb}{0.56, 0.93, 0.56}
\definecolor{customblue}{rgb}{0.2, 0.6, 1.0}
\definecolor{pink}{rgb}{1.0, 0.6, 0.8}

\title{Towards Building General Purpose Embedding Models for Industry 4.0 Agents}
\author{
 \textbf{Christodoulos Constantinides\textsuperscript{1}},
 \textbf{Shuxin Lin\textsuperscript{2}},
 \textbf{Dhaval Patel\textsuperscript{2}}
\\
 \textsuperscript{1}IBM,
 \textsuperscript{2}IBM Research
\\
 \{christodoulos.constantinides@, shuxin.lin@, pateldha@us.\}ibm.com
}
\begin{document}
\maketitle
\begin{abstract}
% In modern industrial settings, maintaining and predicting the health of critical assets is essential for minimizing downtime and preventing costly failures. This paper introduces a framework designed to assist reliability engineers with maintenance-related tasks, inspired by recommender systems. We domain-adapt embedding models to handle different tasks, which are then invoked by an LLM-based recommender agent that serves as the interface between users and the embedded tools. Given a set of tasks expressed in natural language—each associated with queries related to specific assets—we train the embedding models to ``recommend'' relevant items for similar queries on new assets. For instance, a task might involve identifying sensors for a given failure mode and asset, with a query like, ``What sensors are relevant for detecting an eccentric rotor in a wind turbine?'' and candidate items such as ``vibration'', ``speed'' and ``temperature'' sensors. We evaluate the various components of our system and demonstrate its effectiveness in supporting reliability engineers in their day-to-day operations.

In this work we focus on improving language models' understanding for asset maintenance to guide the engineer's decisions and minimize asset downtime. Given a set of tasks expressed in natural language for Industry 4.0\footnote{Industry 4.0 is the integration of advanced technologies and automation into manufacturing and industrial processes to create smart, interconnected, and efficient systems.} domain, each associated with queries related to a specific asset, we want to recommend relevant items and generalize to queries of similar assets. A task may involve identifying relevant sensors given a query about an asset's failure mode. 

Our approach begins with gathering a qualitative, expert-vetted knowledge base to construct nine asset-specific task datasets. To create more contextually informed embeddings, we augment the input tasks using Large Language Models (LLMs), providing concise descriptions of the entities involved in the queries.
This embedding model is then integrated with a Reasoning and Acting agent (ReAct), which serves as a powerful tool for answering complex user queries that require multi-step reasoning, planning, and knowledge inference.

Through ablation studies, we demonstrate that: (a) LLM query augmentation improves the quality of embeddings, (b) Contrastive loss and other methods that avoid in-batch negatives are superior for datasets with queries related to many items, and (c) It is crucial to balance positive and negative in-batch samples. After training and testing on our dataset, we observe a substantial improvement: \textbf{HIT@1 increases by +54.2\%, MAP@100 by +50.1\%, and NDCG@10 by +54.7\%}, averaged across all tasks and models. Additionally, we empirically demonstrate the model's planning and tool invocation capabilities when answering complex questions related to industrial asset maintenance, showcasing its effectiveness in supporting Subject Matter Experts (SMEs) in their day-to-day operations. \textbf{We open-source implementation and experiments.}
\end{abstract}

% We domain-adapt embedding models to handle different tasks, which are then invoked by an LLM-based recommender agent that serves as the interface between users and the embedded tools. Given a set of tasks expressed in natural language—each associated with queries related to specific assets—we train the embedding models to ``recommend'' relevant items for similar queries on new assets. For instance, a task might involve identifying sensors for a given failure mode and asset, with a query like, ``What sensors are relevant for detecting an eccentric rotor in a wind turbine?'' and candidate items such as ``vibration'', ``speed'' and ``temperature'' sensors. We evaluate the various components of our system and demonstrate its effectiveness in supporting reliability engineers in their day-to-day operations.

% Don't go very specific (no reliability engineer)
% section 1. contribution: we identify 9 tasks (people like numbers, be specific). Say the improvement (e.g. average improvement) in numbers
% no need for bullets in 1.1
% Can remove organization text
% 5.2, can make a small paragraph, because I already have the figure
% contribution: extra bullet highlighting the experiments and llm augmentation. 
\section{Introduction}
As Large Language Models (LLMs) and AI agents continue to mature, automating a wide range of tasks is emerging as the next wave of innovation. These tasks range from general activities, such as reading emails, to more domain-specific queries. For example, in industrial settings, a plant operator might seek recommendations for the set-points of critical control variables, while an industrial data scientist could need assistance in identifying key sensor variables for predictive modeling. AI agents, acting as role-playing personas, are increasingly being employed to tackle such tasks by leveraging LLMs and external, domain-specific tools accessed via APIs.
One such example is MDAgent \cite{kim2024mdagents}, a multi-agent system designed for the medical domain to address disease-related queries. While MDAgent helps guide users through predefined options, it relies solely on an LLM’s internal knowledge for decision-making. This limits its effectiveness when dealing with more complex, real-world problems. Therefore, there is a growing need for more flexible, context-sensitive solutions capable of handling intricate, multifaceted decision-making.

\begin{figure*}[h!]
  \centering
  \includegraphics[width=0.8\linewidth]{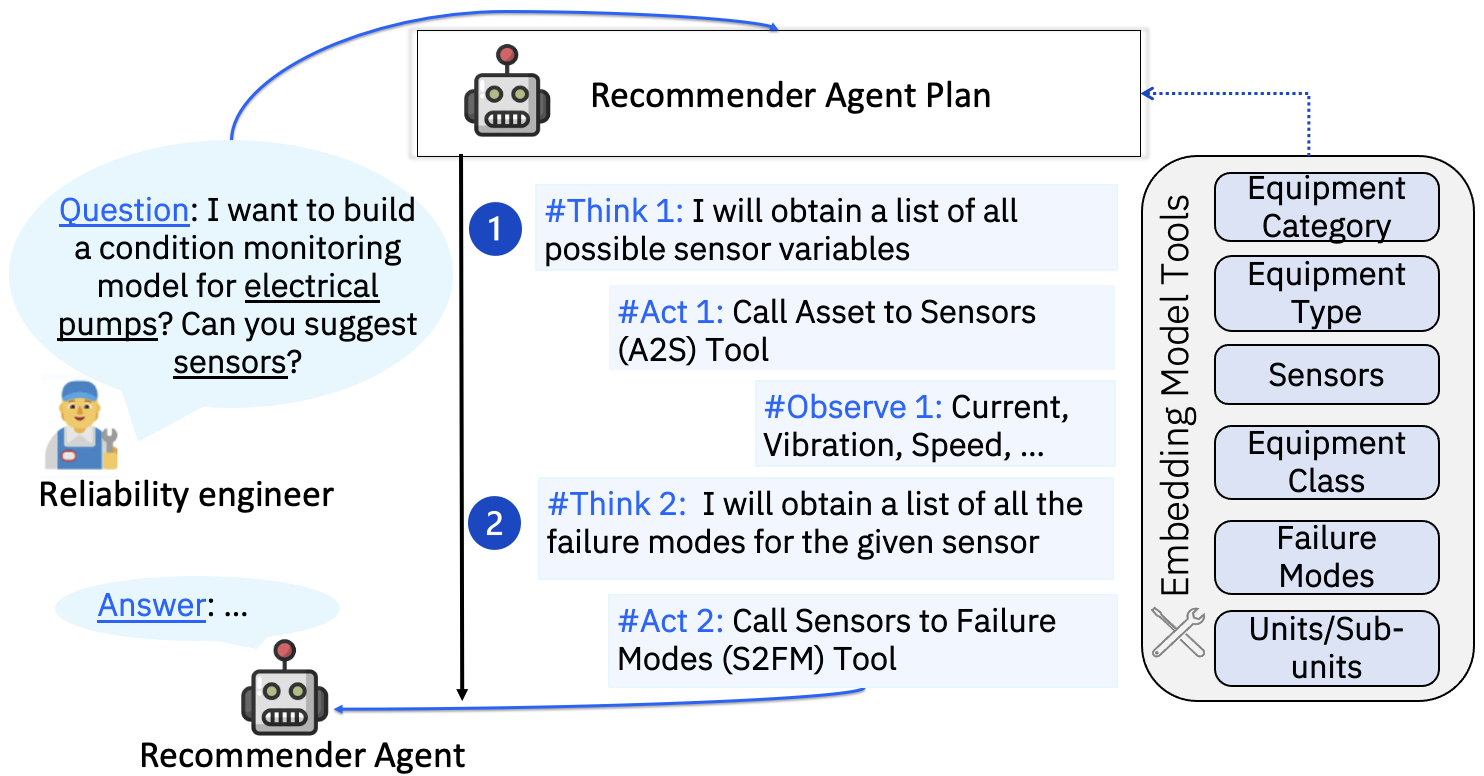}
  \caption{Integration of industrial tools with a reasoning agent.}
  \label{fig:system_diagram}
\end{figure*}

Recommender systems present a promising alternative for assisting users in navigating complex decision spaces. These systems excel at learning from user behavior, item properties, and historical data, providing personalized decision support. Recently, researchers have integrated recommender systems into AI agents via embedding tool, particularly in the entertainment domain \cite{gao2023chatrecinteractiveexplainablellmsaugmented}, to enhance user interaction while addressing challenges related to interactivity and explainability. A key advancement in this area is the development of instruction-driven, domain-specific embeddings, which provide task-specific prompts to the embedding model. This approach improves the relevance of the retrieved information compared to general-purpose embeddings, which may not be as contextually specific. Additionally, leveraging LLMs as teacher models to generate synthetic data has enabled more effective training of domain-specific embeddings \cite{wang2024improvingtextembeddingslarge}. Collectively, these advancements signify the movement toward more efficient, context-aware embedding techniques that can better support complex decision-making in specialized domains.

\subsection{Challenges in Industry 4.0}
Despite the potential of domain-specific embedding techniques as powerful tools within the AI agent framework, their application in Industry 4.0 remains under-explored. Implementing embedders in industrial settings presents several unique challenges.

One key challenge is the accessibility of specialized knowledge. Actionable, fine-grained information is often embedded in technical documents, such as International Standards Organization (ISO) manuals, which are not structured for immediate use and are difficult to access in practice.

A second challenge lies in task-specific instruction. Instruction-tuned embedders depend on well-defined, domain-relevant guidance, yet much of the necessary instructional knowledge in industrial contexts is informal or remains only partially formalized.

A third concern involves asset-related knowledge. Industrial environments comprise a large number of heterogeneous assets, but the amount of information available per asset is often sparse. Developing techniques to augment and cross-leverage knowledge across assets is still an open problem.

To address these limitations, we propose a multi-task, asset-specific fine-tuning strategy that aims to bridge domain gaps and enable more context-aware, robust intelligence within agentic workflows.

Our main contributions are as follows:
\begin{itemize}
    \item We introduce the first embedding models specialized for the industrial domain, designed to assist SMEs with various tasks. We identify nine tasks from ISO documents and train a multi-task embedder (Figure \ref{fig:embedding_training}) to retrieve the final answer rather than generating it. This approach enables easier tracking and improvement of performance. Retrieval metrics such as Accuracy@K and Precision@K are more intuitive compared to text generation evaluation metrics like Perplexity, which measures at the token level. We demonstrate an average increase of ACC@1 by 54\% for our use case.
    \item We demonstrate through ablation studies that popular representation learning methods that use in-batch negatives are prone to false negatives in certain data settings. Additionally, we highlight the importance of maintaining a balance between positive and negative samples within the batch.
    \item We demonstrate how this multi-task embedder can serve as a set of tools (Figure \ref{fig:system_diagram}) invoked by a ReAct agent \cite{yao2022react}, which is capable of planning and reasoning to answer industrial questions, such as ``Which sensors can identify a stalled compressor?''. 
    \item To address the data scarcity issue, we provide the prepared dataset for future research work. 
\end{itemize}

\section{Industrial Multi-Task Embedder}
\label{methodology}
% In this section, we present our multi-task embedder for the industrial domain, with for modeling the the relationships between various entities such as assets, components, sensors, failure modes, and others. 

% Our approach, as outlined in Figure \ref{fig:embedding_training}, leverages tabular information, which are subsequently enriched using LLMs for improved query relevance and answer generation, to train an embedding model.

% This section discusses the problem formulation, followed by an overview of the data sources and tasks collected for the system. The dataset preparation process is then described, along with the augmentation of queries using large language models (LLMs) to enhance the quality of embeddings. The rationale behind the data splitting decisions is also presented, justifying the approach for partitioning the data for training and evaluation. Further, the generation of embeddings is discussed, along with the selection of the loss function for contrastive learning, including a detailed explanation of the reasoning behind this choice to optimize model performance.

\subsection{Foundational Concepts in Industry 4.0}
This section introduces key terminology and foundational concepts relevant to tasks and frameworks in Industry 4.0 applications, particularly in the areas of predictive maintenance, asset management, and sensor-based monitoring. We define six core concepts—referred to here as \textbf{items}—that are central to understanding and implementing industrial AI systems.

\emph{Asset}: A physical resource or piece of equipment used in industrial operations or production processes. Examples include electric generators, transformers, and wind turbines.

\emph{Equipment Class}: A grouping of assets based on shared functional or operational characteristics. For example, ``combustion engines'' form an equipment class that includes subtypes like ``diesel engines'' and ``gas turbines''.

\emph{Equipment Type}: A specific category within an equipment class that characterizes the function or configuration of an asset. A diesel power generator, for instance, is an equipment type under the ``diesel engine'' class.

\emph{Failure Mode}: A particular way in which an asset can fail, including forms of degradation or malfunction. An example is ``insulation deterioration'' in electric motors.

\emph{Sensor}: A device used to measure or monitor a physical parameter such as temperature, vibration, pressure, or rotational speed, often for the purpose of detecting abnormal conditions.

\emph{Subunit}: A defined functional component within a larger system, typically responsible for a specific task. An example is the ``fuel feed system'' in a boiler, which manages the delivery of fuel to the combustion chamber.

\subsection{Data Collection}
\label{tasks}
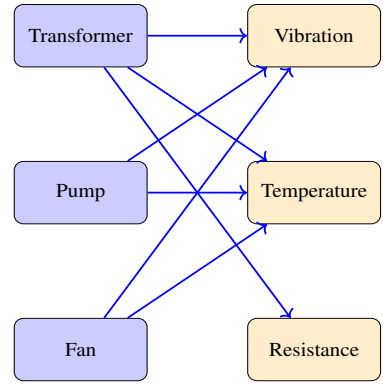
\begin{figure}[h]
    \centering
    \resizebox{5cm}{5cm}{
    \begin{tikzpicture}[node distance=1.5cm, every node/.style={draw, rounded corners, text centered, minimum height=1cm, minimum width=2cm}]
        
        % Set 1 (Assets) - Using fontawesome icons for visual appeal
        % \node[fill=blue!20] (a1) { \faBolt \hspace{0.2cm} Power Transformer};
        % \node[fill=blue!20, below=of a1] (a2) { \faBolt\hspace{0.2cm} Pump};
        % \node[fill=blue!20, below=of a2] (a3) { \faFan \hspace{0.2cm} Fan};
        \node[fill=blue!20] (a1) {\small Transformer};
        \node[fill=blue!20, below=of a1] (a2) {\small Pump};
        \node[fill=blue!20, below=of a2] (a3) {\small Fan};

        % Set 2 (Sensors) - Using icons
        \node[fill=orange!20, right=of a1] (s1) {\small Vibration};
        \node[fill=orange!20, below=of s1] (s2) {\small Temperature};
        \node[fill=orange!20, below=of s2] (s3) {\small Resistance};
        
        % \node[fill=orange!20, right=of a1] (s1) { \faCloud \hspace{0.2cm} Vibration};
        % \node[fill=orange!20, below=of s1] (s2) { \faCloud \hspace{0.2cm} Temperature};
        % \node[fill=orange!20, below=of s2] (s3) { \faLeaf \hspace{0.2cm} Resistance};

        % Drawing edges (connections between assets and sensors)
        \draw[thick, ->, color=blue] (a1) -- (s1);
        \draw[thick, ->, color=blue] (a1) -- (s2);
        \draw[thick, ->, color=blue] (a1) -- (s3);
        \draw[thick, ->, color=blue] (a2) -- (s1);
        \draw[thick, ->, color=blue] (a2) -- (s2);
        % \draw[thick, ->, color=blue] (a2) -- (s3);
        \draw[thick, ->, color=blue] (a3) -- (s1);
        \draw[thick, ->, color=blue] (a3) -- (s2);
        % \draw[thick, ->, color=blue] (a3) -- (s3);

    \end{tikzpicture}
    }
    \caption{Bipartite graph showing relationship ``is monitored'' between assets and sensors.}
    \label{fig:bipartite_graph}
\end{figure}

We gathered documents from ISO that contain related information on maintaining industrial assets \cite{ISO17359_2018}, \cite{ISO14224:2016}. The data were originally in different tabular forms. We prepared a sample bipartite graph to illustrate the mapping between various assets (e.g., motors, machines) and their corresponding sensors (e.g., temperature, environmental) as shown in Figure \ref{fig:bipartite_graph}. The assets are represented as purple color nodes, while the sensors are shown as yellow color nodes, with directed edges indicating the relationships between them. Given an Asset 1, the nodes it is connected to represent \textbf{positive documents}, while the missing edges indicate \textbf{negative documents}.

% TODO: make font smaller

\subsubsection{Tasks Definition}
\label{task-definition}

We identify \textbf{nine distinct tasks} that are integral to our system, each targeting a specific aspect of industrial asset understanding. The underlying data used for generating tasks is extract from structured tabular formats as discussed in Section \ref{tasks}. 

\begin{itemize}
  \item \textbf{Asset to Sensors (A2S):} Identify relevant sensors that can monitor the condition of a given asset.
  \item \textbf{Component to Failure Mode (C2FM):} List possible failure modes associated with a given asset component.
  \item \textbf{Equipment Class Type to Category (E2CAT):} Determine the category corresponding to a given equipment class type.
  \item \textbf{Equipment Category to Class Type (E2CLT):} Identify class types that fall under a given equipment category.
  \item \textbf{Equipment Unit to Subunit (EU2SU):} Identify subunits belonging to a given equipment unit.
  \item \textbf{Failure Mode to Class (FM2CLS):} Classify a given failure mode under its appropriate failure mode class.
  \item \textbf{Failure Mode to Components (FM2CMP):} Identify components associated with a given failure mode and asset.
  \item \textbf{Failure Mode to Sensor (FM2S):} Identify sensors capable of detecting a given failure mode in a specific asset.
  \item \textbf{Sensor to Failure Mode (S2FM):} List failure modes that can be detected by a given sensor for a specific asset.
\end{itemize}

For instance, the task \emph{Failure Mode to Sensor} (FM2S) focuses on identifying appropriate sensors for detecting a given failure mode in a specific asset. We have provided an example prompt and expected response for the FM2S task in Figure~\ref{prompt:example-task}, and representative examples for each task are included in Appendix Table~\ref{tab:task-examples}.

\begin{figure}[h!]
\raggedright
\resizebox{.48\textwidth}{!}{

\begin{tcolorbox}[colframe=black, colback=white, sharp corners, boxrule=0.8mm, width=11cm]
\begin{tcolorbox}[colback=lightgreen!30!white, boxrule=0mm, colframe=lightgreen!50!black, width=10cm, sharp corners, halign=left]
\textbf{Instruct:} What sensors can be applied to detect a fault in an asset and its category?
\end{tcolorbox}
\begin{tcolorbox}[colback=lightgreen!30!white, boxrule=0mm, colframe=lightgreen!50!black, width=10cm, sharp corners, halign=left]
\textbf{Query:} \underline{Asset}: electric motor, \underline{Category}: electric, \underline{Fault:} stator windings fault
\end{tcolorbox}
\begin{tcolorbox}[colback=customblue!10!white, boxrule=0mm, colframe=customblue!50!black, width=10cm, sharp corners, halign=left]
\textbf{Asset description:} Converts electrical energy into mechanical energy to power various industrial machinery.
\end{tcolorbox}
\begin{tcolorbox}[colback=customblue!10!white, boxrule=0mm, colframe=customblue!50!black, width=10cm, sharp corners, halign=left]
\textbf{Fault description:} A Stator windings fault is a type of industrial failure mode where the electrical windings in the stator of an electric motor or generator become damaged or degraded, often due to overheating, insulation breakdown, or physical stress, leading to reduced performance, efficiency, or complete failure of the equipment.
\end{tcolorbox}
\begin{tcolorbox}[colback=pink!10!white, boxrule=0mm, colframe=pink!50!black, width=10cm, sharp corners, halign=left]
\textbf{Sensors:} Current, Vibration, Temperature, Axial Flux, Cooling Gas
\end{tcolorbox}
\noindent
\colorbox{lightgreen!30!white}{Input}, \colorbox{customblue!10!white}{LLM augmented}, \colorbox{pink!10!white}{Ground truth}
\end{tcolorbox}
}
\caption{Example query with LLM augmentation and answer for the FM2S task.}
\label{prompt:example-task}
\end{figure}

\subsubsection{Task Distribution}
From a quantitative perspective, Table \ref{tab:data_dist} presents a distribution across various tasks, focusing on the number of queries, the number of items, and the average number of related items per query. There is significant variability in both the number of items and the average related items per query across tasks, indicating diverse levels of complexity in task structure.

\begin{table}[h!]
    \centering
    \begin{tabular}{|>{\columncolor{gray!20}}p{1.4cm}|>{\columncolor{gray!10}}p{1.3cm}|>{\columncolor{gray!10}}p{1.2cm}|>{\columncolor{gray!10}}p{2cm}|}
        \hline
        \rowcolor{gray!40} 
        \textbf{Task} & \textbf{Query Count} & \textbf{Item Count} & \makecell{\textbf{Avg. Items} \\ \textbf{per Query}} \\
        \hline
        A2S   & 10   & 53   & 12.6 \\ \hline
        C2FM  & 44   & 6    & 1.0  \\ \hline
        E2CAT & 10   & 107  & 10.7 \\ \hline
        E2CLT & 42   & 156  & 4.5  \\ \hline
        EU2SU  & 43   & 1191 & 33.1 \\ \hline
        FM2CLS & 140  & 62   & 1.0  \\ \hline
        FM2CMP & 254  & 44   & 2.7  \\ \hline
        FM2S  & 111  & 53   & 4.5  \\ \hline
        S2FM  & 485  & 55   & 1.0  \\ 
        \hline
    \end{tabular}
    \caption{Distribution of raw data points across tasks.}
    \label{tab:data_dist}
\end{table}

% We prepared a sample bipartite graph to illustrate the mapping between various assets (e.g., motors, machines) and their corresponding sensors (e.g., temperature, environmental) as shown in Figure \ref{fig:bipartite_graph}. The assets are represented as purple color nodes, while the sensors are shown as yellow color nodes, with directed edges indicating the relationships between them. Given an Asset 1, the nodes it is connected to represent \textbf{positive documents}, while the missing edges indicate \textbf{negative documents}.

% For example, in the \textbf{A2S} (Asset to Sensor) task, 10 assets (e.g., electrical motors) are mapped to 53 sensors. The \textbf{S2FM} task has the highest number of queries (485), whereas \textbf{A2S} has the fewest (10). Similarly, \textbf{EU2SU} has the most items (1191), and \textbf{C2FM} has the fewest (6). This shows that some tasks are query-intensive with fewer items, while others involve fewer queries but more items per query. There is significant variability in both the number of items and the average related items per query across tasks, indicating diverse levels of complexity in task structure.

\begin{figure*}[h]
  \centering
\includegraphics[width=0.65\linewidth]{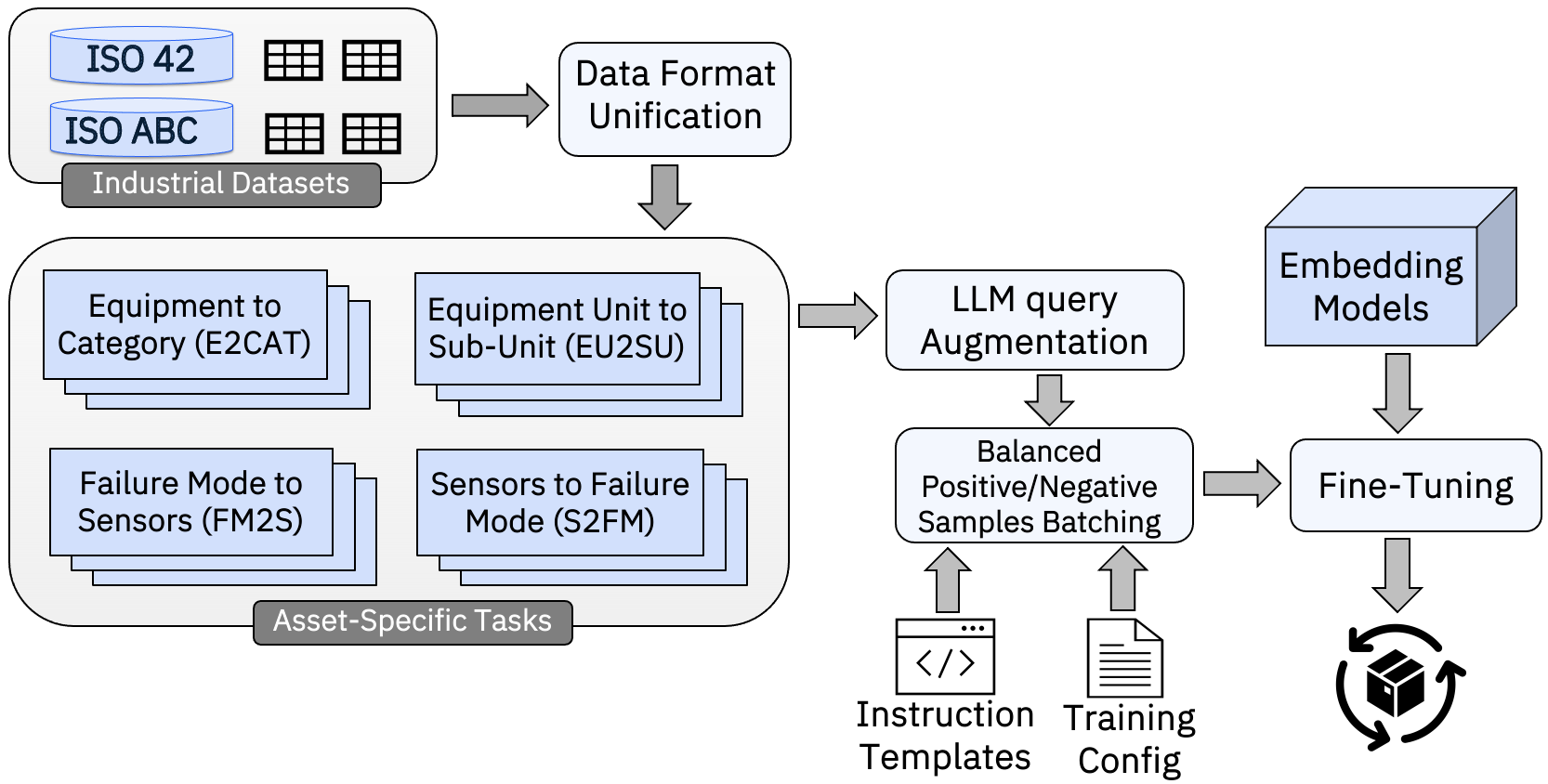}
  \caption{Overall training flow for the tools.}
  \label{fig:embedding_training}
\end{figure*}

\section{Problem Formulation and Solution}
In this section, we present our multi-task embedder for the industrial domain, with for modeling the the relationships between various entities such as assets, components, sensors, failure modes, and others. Our approach, as outlined in Figure \ref{fig:embedding_training}, leverages tabular information, which are subsequently enriched using LLMs for improved query relevance and answer generation, to train an embedding model.

\subsection{Problem Formulation}
\label{problem-formulation}

Given a set of tasks \( T \), where each task \( T_i \in T \) consists of a set of queries \( Q \), and each query \( q \in Q \) is associated with a set of relevant items \( I \), the objective is to model the relationships between queries and their corresponding items. Each task involves queries grounded in various assets \( A \), and semantically similar items may share similar relationships across tasks.

For example, consider a task \( T_i \) focused on identifying relevant sensors. Each query includes an asset from \( A \) (e.g., an \textit{electric generator}) and a failure mode (e.g., \textit{misalignment}). The relevant items—drawn from the set \( I \)—are sensors (e.g., \textit{vibration sensor}) capable of detecting that failure mode for the given asset.

The type of relevant items varies by task: depending on the context, items may represent \textit{sensors}, \textit{failure modes}, \textit{components}, or \textit{sub-components}.

During training, the model observes queries from each task involving a subset of assets and their associated items. The goal is to generalize these relationships, enabling the system to infer relevant items for unseen queries involving different assets.

To this end, we learn a function \( f \) that maps both queries and items into a shared latent space \( \mathbb{R}^d \), such that embeddings of relevant query-item pairs are positioned closer together. At inference time, the system retrieves the top-\( k \) most relevant items for a given query based on proximity in this latent space.

\subsection{Instruction template}
\label{dataset-prep}
Given a relevant query-document pair \((q^+, d^+)\) sampled from Table~\ref{tab:data_dist}, we construct training examples using an instruction-tuning template. Each input is formatted as:
\begin{align*}
q^+_{\text{inst}} &= \texttt{Instruct:}~\{\textit{instruction}\}~\texttt{. Query:}~q^+, \\
o^+_{\text{inst}} &= \{\textit{output-tag}\}~\texttt{:}~d^+.
\end{align*}

where \{\textit{instruction}\} is a natural-language sentence describing the task, and \(q^+\) is the original query. For each task, we generate multiple paraphrased versions of the instruction using a large language model (LLM) to promote generalization.

Figure~\ref{prompt:example-task} illustrates a complete prompt, showing the \texttt{Instruct} and \texttt{Query} fields for a task involving electric motors. Table~\ref{tab:task-examples} in Appendix provides one representative instruction and query pair per task. 

\subsection{LLM query augmentation}
\label{llm-query-aug}
From the tabular data in the ISO documents, the information is not very descriptive to build a good model. For instance, for the sensor to failure mode task the query contains only the asset name, and its sensor. This makes it difficult to learn semantics for this asset and failure mode generalize on new assets. For this reason, we augment the query using an LLM with a one sentence description of its entities. 
We augment with a probability $p$, which acts a type of dropout to avoid overfitting. Figure \ref{prompt:example-task} provides an example augmentation (light blue color). We provide further impacts of this design decision on the ablation studies in Section \ref{effects-p-desc}.

Specifically, we prompt the LLM with instructions such as: ``Provide a one sentence description for the equipment category,'' ``Provide a one sentence description for the equipment type,'' ``Provide a one sentence description for the industrial component,'' and ``Provide a one sentence description for the industrial failure mode.'' These concise descriptions help encode semantic context and enable the embedding model to better capture task-relevant relationships.

\subsection{Data splitting}
\label{data-splitting}
To make the experiment unbiased and more realistic, we split the train/validation/test set queries by assets for the tasks that is possible, otherwise we split randomly. 
% This ensures that we consider a realistic scenario, where we want to see how good our model is on new assets. 

\subsection{Query and Document Embeddings}
\label{doc-embeddings}

Depending on the model used, we employ either the \textit{mean pooling} or \textit{last token pooling} strategy—two common techniques for generating fixed-size embeddings from variable-length token sequences.

\paragraph{Last Token Pooling.}
Given a relevant query \( q^+ \) and document \( d^+ \), we concatenate them with an \texttt{[EOS]} token. The resulting sequence is passed through the transformer model \( f \), and we extract the embedding corresponding to the final \texttt{[EOS]} token from the last hidden layer.

\paragraph{Mean Pooling.}
Let \( T_1, \ldots, T_n \) be the tokens of the instruction-formatted query \( q^+_{\text{inst}} \). These tokens are input to the model \( f \), and we compute the average of the last-layer activations to obtain the final embedding. The same procedure is applied to the instruction-formatted document \( d^+_{\text{inst}} \).

\subsection{Loss Function and Batching}
\label{contrastive-learning}

To learn the embedding function \( f \), we use a contrastive learning framework~\cite{hadsell2006dimensionality}. Given a labeled triplet \( \langle q, d, l \rangle \)—where \( l = 1 \) indicates a relevant query-document pair and \( l = 0 \) indicates a negative pair—we minimize the following margin-based loss:
\[
\mathbb{L} = l \cdot d(q, d)^2 + (1 - l) \cdot \max(\epsilon - d(q, d), 0)^2
\]
Here, \( d(q, d) \) denotes the Euclidean distance between the query and document embeddings, and \( \epsilon \) is a margin parameter.

During training, for each task, we use all available positive \((q^+, d^+)\) and negative \((q^+, d^-)\) pairs. Since the number of negative pairs significantly exceeds the number of positive ones, we balance each training batch to include an equal number of positives and negatives. Additional analysis on this design choice is presented in the ablation study (Section~\ref{varying-positives-negatives}).

\begin{figure}[h]
    \centering
    \includegraphics[width=1\linewidth]{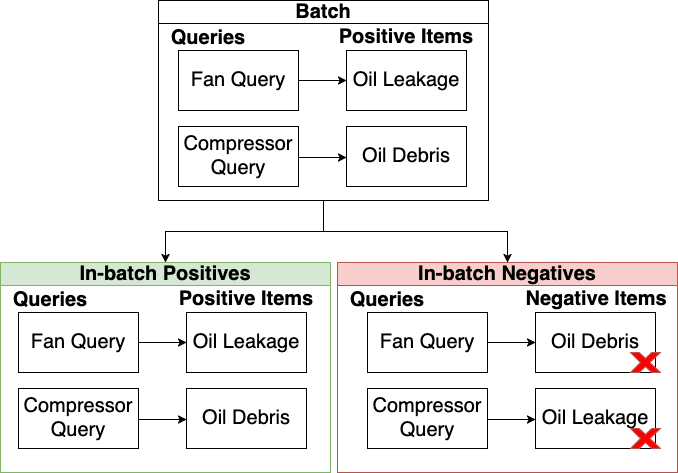}
    \caption{An example of how in-batch negatives is prone to false negatives. Both Fan and Compressor are related to both Oil Leakage and Oil Debris. In the selected batch, the positive items will be used as negatives and this will result in in-batch false negatives. In our dataset there is a high chance of this happening, since for each query there is a high number of related items when compared to the total unique items.}
    \label{fig:in-batch-negatives}
\end{figure}

We avoid loss functions that rely on in-batch negatives due to the high risk of false negatives—caused by the relatively small item set and the fact that each query can be associated with multiple valid items (see Figure~\ref{fig:in-batch-negatives}). In Section~\ref{effects-loss-func}, we compare the performance of our selected loss function with an in-batch contrastive loss variant.

\section{Experimental Setup}
\label{experimental-setup}
In this section we present the comparison before and after fine-tuning several embedding models on the 9 industrial tasks we prepared. 

\subsection{Embedding models}
We compare different models in retrieving the correct answer from a set of candidate items. The models used are: BERT \cite{kenton2019bert}, MPNet \cite{song2020mpnet}, BGE-large-v1.5 \cite{bge_embedding}, gte-Qwen2-7B-instruct \cite{li2023towards}, and E5-Mistral-7B-Instruct \cite{wang2023improving} which have varying sizes. We also compare against BM25 retriever \cite{robertson2009probabilistic}, which is a bag-of-words retrieval function.

% TODO: add some comments about best model etc
% \clearpage

\subsection{Training settings}
For all the models we used 4 A100 80GB GPUs with a training batch size of 32 per device. We do a full fine-tuning for BERT, MPNet, and BGE-large-v1.5, while for gte-Qwen2-7B-instruct and E5-Mistral-7B-Instruct we use LoRA \cite{hu2021lora} to fit them into the memory. We used mean pooling for all the models to generate the embeddings, apart from Qwen-7B for which we used last token pooling which is how it was trained and has better results. We train a single model on all the different tasks for 3 epochs. We applied query augmentation as described in section \ref{llm-query-aug} with 50\% probability using Llama-3.1-70B-Instruct.
% TODO: show an in-context example.

\begin{figure}[h]
    \centering
    \includegraphics[width=0.9\linewidth]{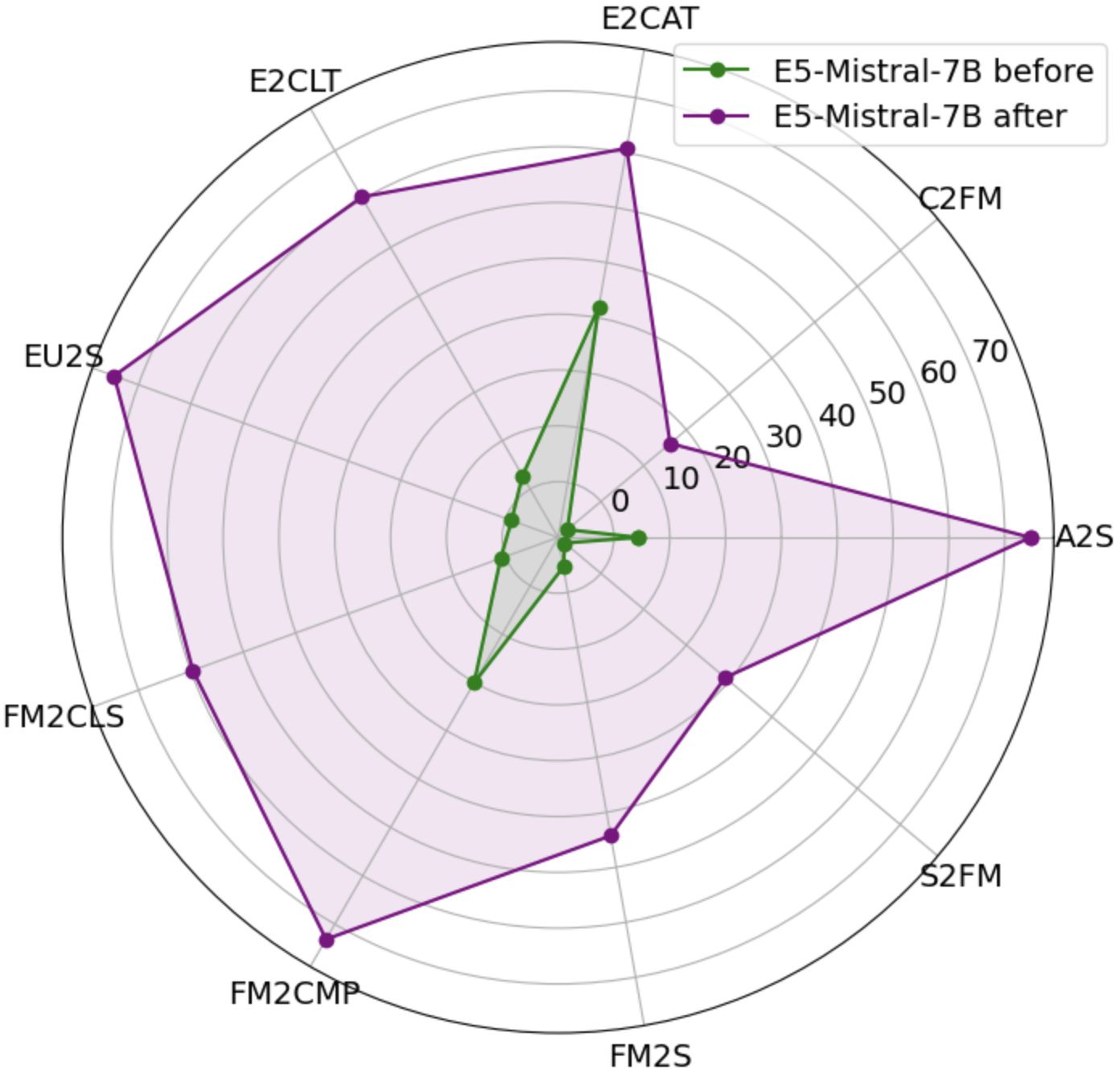}
    \caption{MAP@100 before and after finetuning E5 Mistral 7B for each task.}
    \label{fig:radar-before-after}
\end{figure}

\subsection{Embedding models fine-tuning results}
\begin{figure}
    \centering
    \includegraphics[width=1\linewidth]{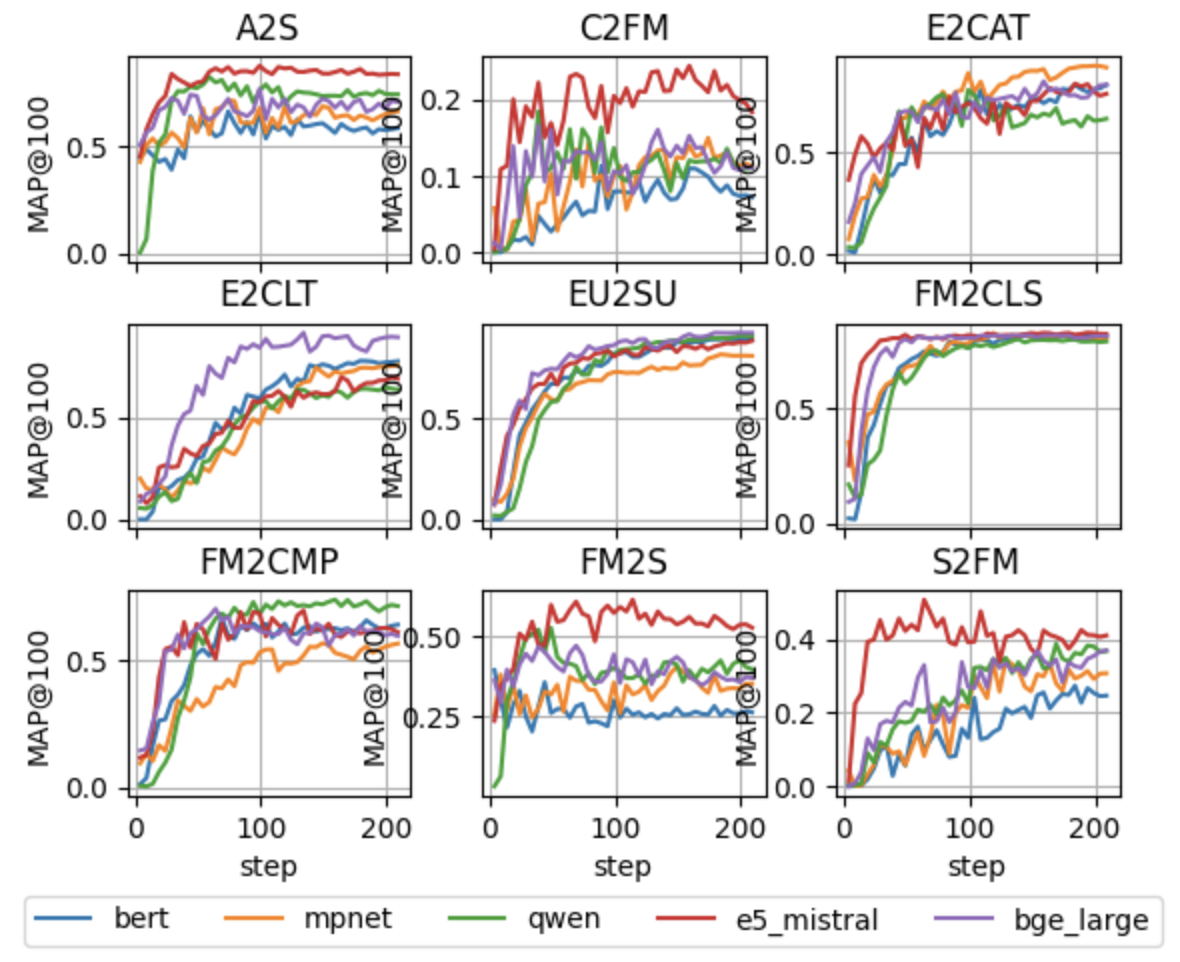}
    \caption{Validation performance during training for every task (MAP@100).}
    \label{fig:val-perf}
\end{figure}

Figure \ref{fig:radar-before-after} provides a result using fine-tuned model for all the task. We observe consistent performance improvements across all tasks after fine-tuning. As shown in Figure~\ref{fig:val-perf}, MAP@100 increases steadily on the validation set during training, demonstrating the effectiveness of our fine-tuning approach. The best-performing model varies by task, highlighting the diverse nature of the task set. Even prior to fine-tuning, the E5 Mistral model demonstrates strong zero-shot capabilities on several tasks, suggesting that it already captures some aspects of instruction semantics. A detailed breakdown of task-level performance before and after fine-tuning is provided in Appendix Table~\ref{table:embeddings-results}.

\section{Ablation study}
\label{ablation-study}

\subsection{Effects of adding LLM-generated description}
\label{effects-p-desc}
\begin{figure}[h!]
    \centering
    \includegraphics[width=1\linewidth]{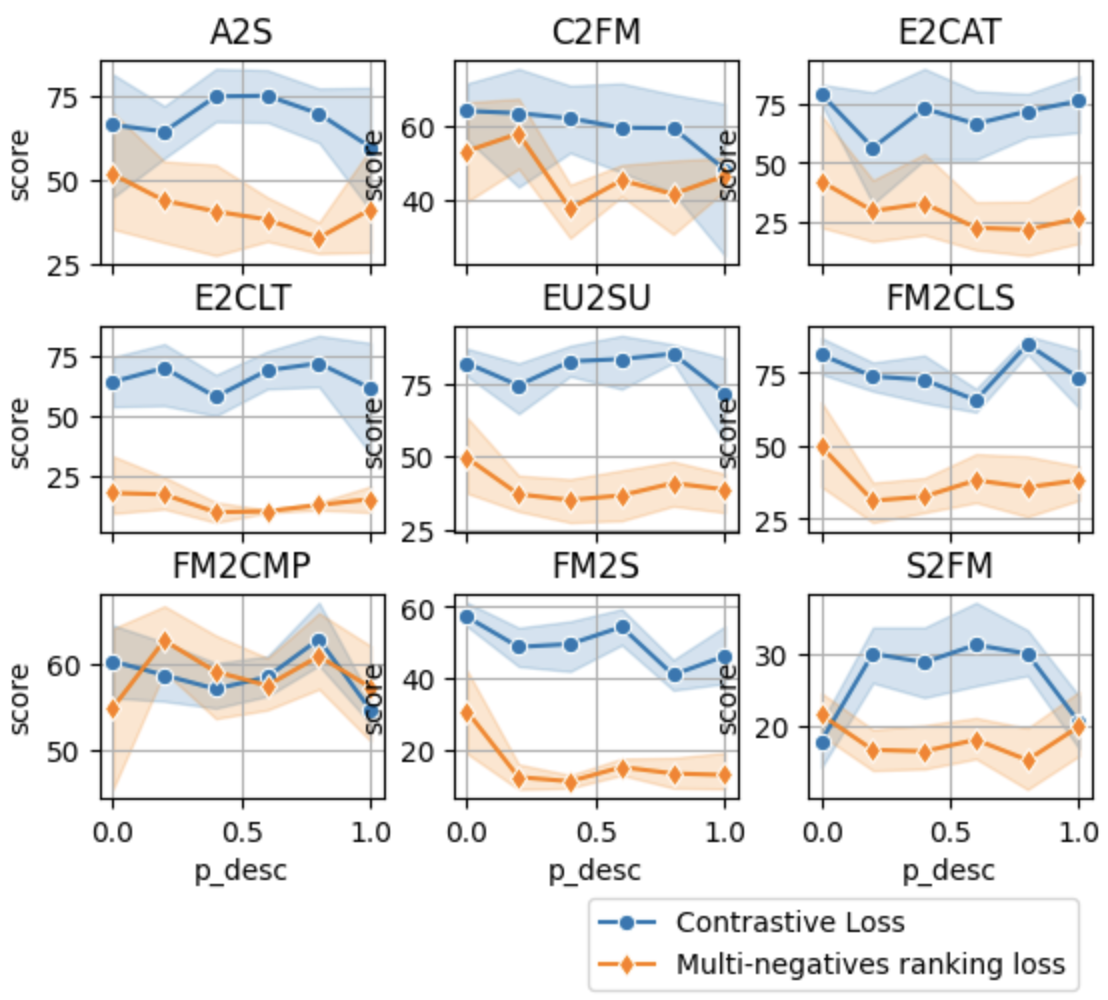}
    \caption{Loss comparison and effects of varying probability of augmenting the query using an LLM on the model's performance (MAP@100) per task.}
    \label{fig:varying-pdesc}
\end{figure}

We investigate what are the effects of adding LLM-generated description on the model's performance. We vary the probability of adding one sentence of LLM-generated description from 0 to 1 with 0.2 strides and see the effects on the test performance for each task (Figure \ref{fig:varying-pdesc}). 
For this experiment we fixed the model to Mistral E5. For some tasks we can see performance boost (A2S, E2CLT, S2FM). It is notable that augmenting the query with a single sentence boosts performance when the probability of augmenting is not 0 or 1. This can be thought as some type of dropout. We believe that the rest of the tasks could also be benefited if the dataset was larger with more queries.

\begin{figure}[h]
    \centering
    \includegraphics[width=1.0\linewidth]{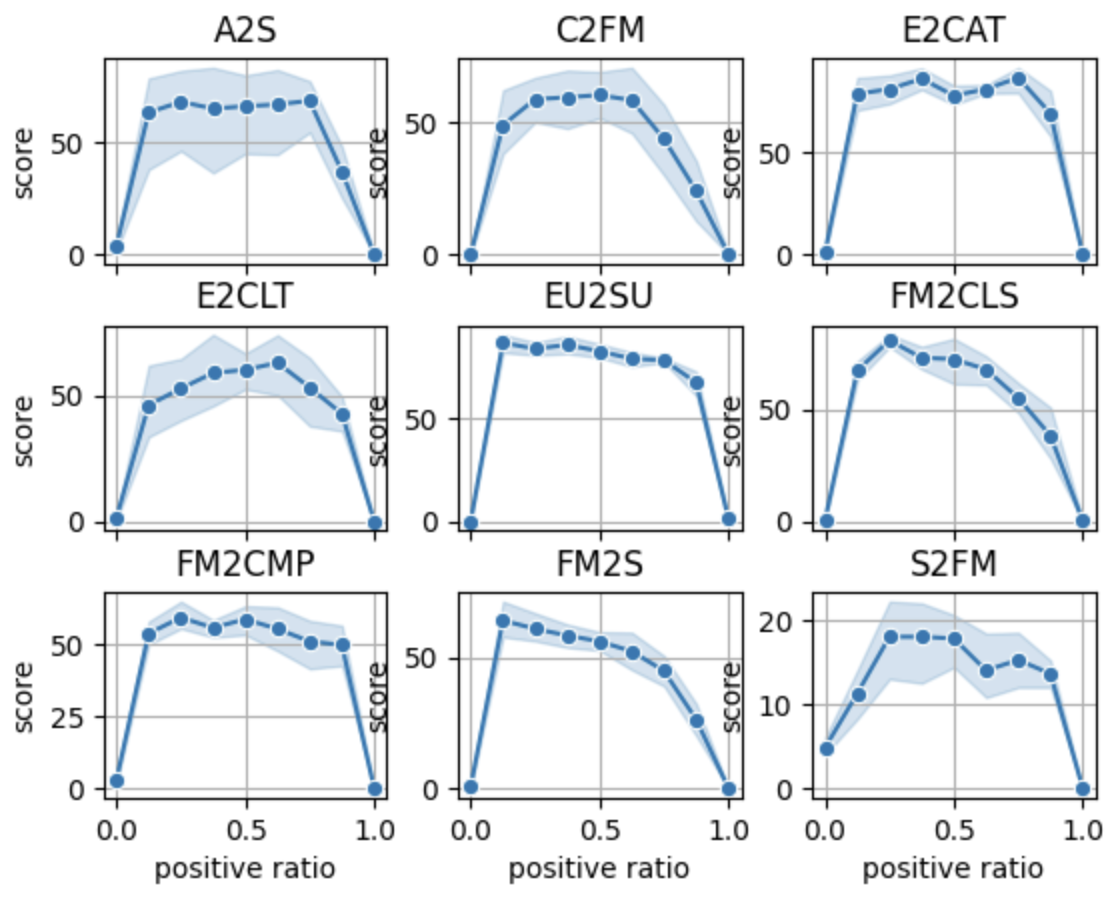}
    \caption{Effects of varying in-batch positives to negatives ratio per task (MAP@100).}
    \label{fig:varying-positive-ratio}
\end{figure}
\subsection{Effects of loss function}
\label{effects-loss-func}
We compare contrastive loss that we adopted in our system (Section \ref{contrastive-learning}) against multi-negatives ranking loss \cite{henderson2017efficient}. One big difference between the two losses, is that multi-negatives ranking loss uses in-batch negatives. The loss function is defined as: 
\[\mathbb{L} = -\log \frac{\phi(q^{+}_{\text{inst}}, d^{+})}{\phi(q^{+}_{\text{inst}}, d^{+}) + \sum\limits_{n_i \in \mathbb{N}} \phi(q^{+}_{\text{inst}}, n_i)}\]
where $\mathbb{N}$ denotes the set of all in-batch negatives, and $\phi(\cdot)$ is the cosine similarity. For this loss function we only provided the query and positive examples, and for the $i_{th}$ query $q_i$ in the batch, all the positives from the rest of the in-batch samples were used as in-batch negatives $d_j$ where $i \neq j$. Figure \ref{fig:varying-positive-ratio} performance of contrastive against multi-negatives ranking loss. The results indicate that the in-batch negatives hurt the performance due to the high chance of false negatives.
\subsection{Effects of in-batch positives to negatives samples ratio}
\label{varying-positives-negatives}
Using Contrastive Loss, we vary the ratio of in-batch positives to negatives and study its impact. The batch size is fixed to 32, and the ratio is varied from 0 to 1 with 0.125 strides. The experiment is repeated 5 times with a different random seed. We chose to not apply any augmentation on the queries to avoid interference with this experiment. Figure \ref{fig:varying-positive-ratio} shows MAP@100 broken down by task for varying positives to negatives ratio. When the batch consists of only positive or only negative samples, the performance is very poor. Overall, a batch with balanced positives and negatives is performing the best.
% \section{LLM recommender application}
% \begin{figure*}[h]
%     \centering
%     \includegraphics[width=12cm, height=8cm]{overall_flow.png}
%     \caption{Integration of the industrial embedding tools with an LLM chat recommender system which serves as the interface between users and the tools. The recommendation agent constructs a plan based on the user's query and invokes the appropriate tools to execute it.}
%     \label{fig:overall_flow}
% \end{figure*}

\begin{figure}[t!]
    \centering
    \includegraphics[width=1\linewidth]{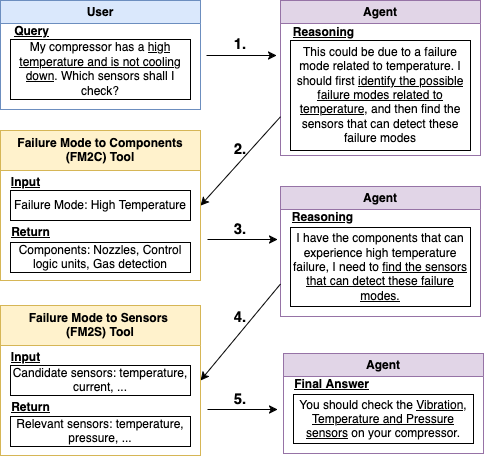}
    \caption{Interaction between user and agent.}
    \label{fig:user-agent-trajectory}
\end{figure}

\section{Case Study}
\label{case-study}
We present a real case study on how the agent interacts with our tools for a given user query. We do a qualitative analysis on the planning of the agent to solve industrial related tasks given our industrial tools.
\subsection{Setup}
We wrap our multi-task, domain-adapted, embedding models as langchain\footnote{https://www.langchain.com/} tools, which accept different arguments. We provide a description for each tool, its arguments, what they return, the data types, and state which arguments are optional. In some of the tools we make calls to an LLM to get a one sentence description of the input entities (e.g. sensor, failure mode) which can help for more meaningful embeddings. We also provide an optional argument for candidate items in case the agent has some candidate items which it wants to pass to the tool to keep the most relevant.   
We provide as context 3 examples to the agent on how to call the tools to solve a simple question. We purposely use examples that are unrelated with each other, so that the agent can brainstorm on new ideas and not follow the same trajectory as the examples in context.

\subsection{Interaction}
Figure \ref{fig:user-agent-trajectory} presents the interaction between user, agent, and our designed industrial tools described in section \ref{methodology}. The reliability engineer is querying the system that he is \textbf{observing high temperature in their compressor}. Then the agent constructs a plan and invokes several of our tools to finally return the relevant sensors, along with generated information around the affected components.
\begin{enumerate}
     \item The reliability engineer is observing high temperature in his compressor.
     \item The agent invokes the Failure Mode to Component (FM2C) tool with the failure mode as argument to discover which components can experience this failure, and goes on and on.
%     \item The agent receives the components from the tool and generates a list of candidate sensors that it thinks they could be effective in capturing the given failure mode.
%     \item The agent invokes the failure mode to sensor tool (FM2S) and provides the information from the previous steps as arguments.
%     \item The agent returns the relevant sensors to the reliability engineer.
\end{enumerate}

\subsection{Discussion}
The agent seems to follow an interesting pattern. It first finds the components that can experience the given fault with tool calling. Then, given all the components, it provides some candidate sensors that it thinks are used to monitor these components (probably from prior knowledge) and asks the LLM to narrow them down to only the relevant ones, getting 2 out of 3 correct in the final response. One thing that is notable is that none of these tools were given as in-context examples to the model, so these decisions from the model seem rather unbiased. 

\section{Conclusion}
In this work we present an innovative framework for building domain adapted tools for multiple tasks starting from tabular data. These tools consist of an embedding model which receives a query, augments it using an LLM and retrieves the answer closest in the embedding space. The experiments quantitatively confirm the effectiveness of our approach after fine-tuning different models, and the ablation studies corroborate our design choices. We then provide a case-study to show how the ReAct agent interacts with these tools which demonstrates an out-of-the-box thinking from the agent. Our framework is easy to follow, provides an easy performance tracking and improvement through retrieval, and can be replicated to other domains.  

\section{Limitations}
We currently fix k when retrieving top-k relevant items in the ReAct agent, which may not match the actual number of relevant results. In the future, we plan to adopt dynamic thresholds (e.g., distance-based or cross-encoder filtering) to improve precision and recall.

Another limitation is that the LLM-generated entity descriptions (Section~\ref{llm-query-aug}) can hallucinate details. We manually removed such cases in this version, but future work will automate this process. We plan to explore attention maps \cite{rateike2023weakly, sriramanan2024llm}, activation analysis \cite{yehuda2024interrogatellm}, and perplexity-based methods \cite{sriramanan2024llm} to better understand and reduce hallucinations and their impact on embedding quality.

\bibliography{acl}

\begin{thebibliography}{21}
\providecommand{\natexlab}[1]{#1}

\bibitem[{Constantinides et~al.(2025)Constantinides, Patel, Lin, Guerrero,
  Patil, and Kalagnanam}]{constantinides2025failuresensoriq}
Christodoulos Constantinides, Dhaval Patel, Shuxin Lin, Claudio Guerrero,
  Sunil~Dagajirao Patil, and Jayant Kalagnanam. 2025.
\newblock Failuresensoriq: A multi-choice qa dataset for understanding sensor
  relationships and failure modes.
\newblock \emph{arXiv preprint arXiv:2506.03278}.

\bibitem[{Gao et~al.(2023)Gao, Sheng, Xiang, Xiong, Wang, and
  Zhang}]{gao2023chatrecinteractiveexplainablellmsaugmented}
Yunfan Gao, Tao Sheng, Youlin Xiang, Yun Xiong, Haofen Wang, and Jiawei Zhang.
  2023.
\newblock \href {https://arxiv.org/abs/2303.14524} {Chat-rec: Towards
  interactive and explainable llms-augmented recommender system}.
\newblock \emph{Preprint}, arXiv:2303.14524.

\bibitem[{Hadsell et~al.(2006)Hadsell, Chopra, and
  LeCun}]{hadsell2006dimensionality}
Raia Hadsell, Sumit Chopra, and Yann LeCun. 2006.
\newblock Dimensionality reduction by learning an invariant mapping.
\newblock In \emph{2006 IEEE computer society conference on computer vision and
  pattern recognition (CVPR'06)}, volume~2, pages 1735--1742. IEEE.

\bibitem[{Henderson et~al.(2017)Henderson, Al-Rfou, Strope, Sung, Luk{\'a}cs,
  Guo, Kumar, Miklos, and Kurzweil}]{henderson2017efficient}
Matthew Henderson, Rami Al-Rfou, Brian Strope, Yun-Hsuan Sung, L{\'a}szl{\'o}
  Luk{\'a}cs, Ruiqi Guo, Sanjiv Kumar, Balint Miklos, and Ray Kurzweil. 2017.
\newblock Efficient natural language response suggestion for smart reply.
\newblock \emph{arXiv preprint arXiv:1705.00652}.

\bibitem[{Hu et~al.(2021)Hu, Shen, Wallis, Allen-Zhu, Li, Wang, Wang, and
  Chen}]{hu2021lora}
Edward~J Hu, Yelong Shen, Phillip Wallis, Zeyuan Allen-Zhu, Yuanzhi Li, Shean
  Wang, Lu~Wang, and Weizhu Chen. 2021.
\newblock Lora: Low-rank adaptation of large language models.
\newblock \emph{arXiv preprint arXiv:2106.09685}.

\bibitem[{ISO(2016)}]{ISO14224:2016}
ISO. 2016.
\newblock \href {https://www.iso.org/standard/64076.html} {Iso 14224:2016 -
  petroleum, petrochemical and natural gas industries — collection and
  exchange of reliability and maintenance data for equipment}.
\newblock Last reviewed and confirmed in 2022; remains current.

\bibitem[{ISO(2018)}]{ISO17359_2018}
ISO. 2018.
\newblock \href {https://www.iso.org/standard/71194.html} {Condition monitoring
  and diagnostics of machines — general guidelines}.
\newblock Geneva, Switzerland. International Organization for Standardization
  (ISO).
\newblock This publication was last reviewed and confirmed in 2023. Therefore,
  this version remains current.

\bibitem[{Kenton and Toutanova(2019)}]{kenton2019bert}
Jacob Devlin Ming-Wei~Chang Kenton and Lee~Kristina Toutanova. 2019.
\newblock Bert: Pre-training of deep bidirectional transformers for language
  understanding.
\newblock In \emph{Proceedings of naacL-HLT}, volume~1. Minneapolis, Minnesota.

\bibitem[{Kim et~al.(2024)Kim, Park, Jeong, Chan, Xu, McDuff, Lee, Ghassemi,
  Breazeal, and Park}]{kim2024mdagents}
Yubin Kim, Chanwoo Park, Hyewon Jeong, Yik~Siu Chan, Xuhai Xu, Daniel McDuff,
  Hyeonhoon Lee, Marzyeh Ghassemi, Cynthia Breazeal, and Hae~Won Park. 2024.
\newblock Mdagents: An adaptive collaboration of llms for medical
  decision-making.
\newblock In \emph{The Thirty-eighth Annual Conference on Neural Information
  Processing Systems}.

\bibitem[{LangChain(2024{\natexlab{a}})}]{langchain_arxiv_tool}
LangChain. 2024{\natexlab{a}}.
\newblock Langchain arxiv tool integration.
\newblock \url{https://python.langchain.com/docs/integrations/tools/arxiv/}.
\newblock Accessed: 2025-06-07.

\bibitem[{LangChain(2024{\natexlab{b}})}]{langchain_wikipedia_tool}
LangChain. 2024{\natexlab{b}}.
\newblock Langchain wikipedia tool integration.
\newblock
  \url{https://python.langchain.com/docs/integrations/tools/wikipedia/}.
\newblock Accessed: 2025-06-07.

\bibitem[{Li et~al.(2023)Li, Zhang, Zhang, Long, Xie, and
  Zhang}]{li2023towards}
Zehan Li, Xin Zhang, Yanzhao Zhang, Dingkun Long, Pengjun Xie, and Meishan
  Zhang. 2023.
\newblock Towards general text embeddings with multi-stage contrastive
  learning.
\newblock \emph{arXiv preprint arXiv:2308.03281}.

\bibitem[{Rateike et~al.(2023)Rateike, Cintas, Wamburu, Akumu, and
  Speakman}]{rateike2023weakly}
Miriam Rateike, Celia Cintas, John Wamburu, Tanya Akumu, and Skyler Speakman.
  2023.
\newblock Weakly supervised detection of hallucinations in llm activations.
\newblock \emph{arXiv preprint arXiv:2312.02798}.

\bibitem[{Robertson et~al.(2009)Robertson, Zaragoza
  et~al.}]{robertson2009probabilistic}
Stephen Robertson, Hugo Zaragoza, et~al. 2009.
\newblock The probabilistic relevance framework: Bm25 and beyond.
\newblock \emph{Foundations and Trends{\textregistered} in Information
  Retrieval}, 3(4):333--389.

\bibitem[{Song et~al.(2020)Song, Tan, Qin, Lu, and Liu}]{song2020mpnet}
Kaitao Song, Xu~Tan, Tao Qin, Jianfeng Lu, and Tie-Yan Liu. 2020.
\newblock Mpnet: Masked and permuted pre-training for language understanding.
\newblock \emph{Advances in neural information processing systems},
  33:16857--16867.

\bibitem[{Sriramanan et~al.(2024)Sriramanan, Bharti, Sadasivan, Saha,
  Kattakinda, and Feizi}]{sriramanan2024llm}
Gaurang Sriramanan, Siddhant Bharti, Vinu~Sankar Sadasivan, Shoumik Saha,
  Priyatham Kattakinda, and Soheil Feizi. 2024.
\newblock Llm-check: Investigating detection of hallucinations in large
  language models.
\newblock \emph{Advances in Neural Information Processing Systems},
  37:34188--34216.

\bibitem[{Wang et~al.(2023)Wang, Yang, Huang, Yang, Majumder, and
  Wei}]{wang2023improving}
Liang Wang, Nan Yang, Xiaolong Huang, Linjun Yang, Rangan Majumder, and Furu
  Wei. 2023.
\newblock Improving text embeddings with large language models.
\newblock \emph{arXiv preprint arXiv:2401.00368}.

\bibitem[{Wang et~al.(2024)Wang, Yang, Huang, Yang, Majumder, and
  Wei}]{wang2024improvingtextembeddingslarge}
Liang Wang, Nan Yang, Xiaolong Huang, Linjun Yang, Rangan Majumder, and Furu
  Wei. 2024.
\newblock \href {https://arxiv.org/abs/2401.00368} {Improving text embeddings
  with large language models}.
\newblock \emph{Preprint}, arXiv:2401.00368.

\bibitem[{Xiao et~al.(2023)Xiao, Liu, Zhang, and Muennighoff}]{bge_embedding}
Shitao Xiao, Zheng Liu, Peitian Zhang, and Niklas Muennighoff. 2023.
\newblock \href {https://arxiv.org/abs/2309.07597} {C-pack: Packaged resources
  to advance general chinese embedding}.
\newblock \emph{Preprint}, arXiv:2309.07597.

\bibitem[{Yao et~al.(2022)Yao, Zhao, Yu, Du, Shafran, Narasimhan, and
  Cao}]{yao2022react}
Shunyu Yao, Jeffrey Zhao, Dian Yu, Nan Du, Izhak Shafran, Karthik Narasimhan,
  and Yuan Cao. 2022.
\newblock React: Synergizing reasoning and acting in language models.
\newblock \emph{arXiv preprint arXiv:2210.03629}.

\bibitem[{Yehuda et~al.(2024)Yehuda, Malkiel, Barkan, Weill, Ronen, and
  Koenigstein}]{yehuda2024interrogatellm}
Yakir Yehuda, Itzik Malkiel, Oren Barkan, Jonathan Weill, Royi Ronen, and Noam
  Koenigstein. 2024.
\newblock Interrogatellm: Zero-resource hallucination detection in
  llm-generated answers.
\newblock \emph{arXiv preprint arXiv:2403.02889}.

\end{thebibliography}

\appendix

%\section{Appendix}
\clearpage

\section{Experiments Details and Results}
In this section, we present details for each task including example queries, LLM augmentation, and answer. Table~\ref{tab:task-examples} presents representative examples for each task type covered in our framework. For each task, we show the input query, the augmentation applied by the LLM (if any), and the corresponding answer. Tasks span a range of reliability and maintenance reasoning types—from sensor selection (A2S, FM2S) to failure mode identification (C2FM, FM2CMP, FM2CLS) and equipment/component mapping (E2CAT, E2CLT, EU2SU). Where applicable, LLMs are prompted to enhance the query context with domain-specific descriptions to improve relevance and answer quality. In some cases (marked with an asterisk), no augmentation is applied as the original query contained sufficient detail.
  
\begin{figure*}
    \centering
    \includegraphics[width=0.8\linewidth]{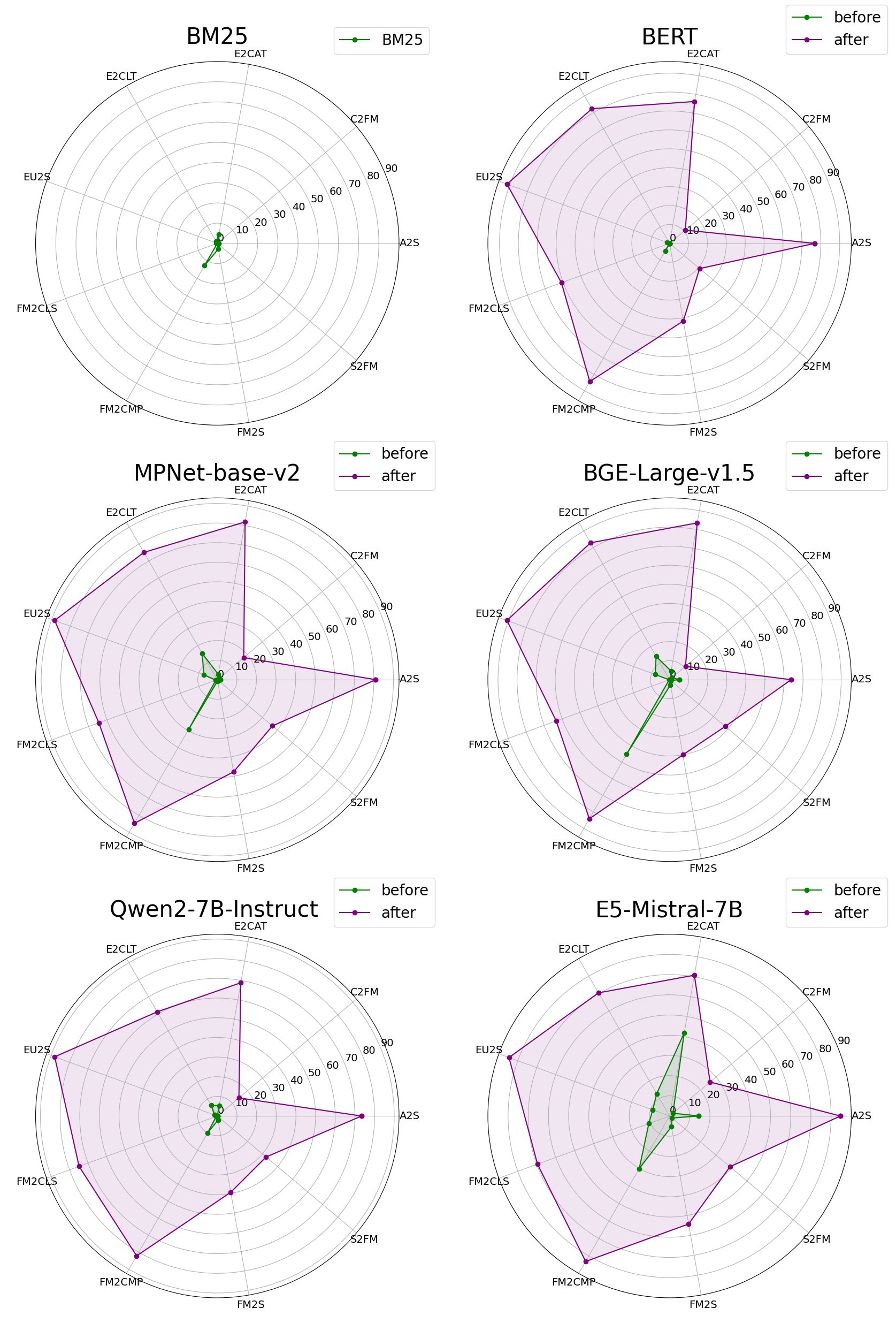}
    \caption{MAP@100 retrieval performance before and after fine-tuning each model and task.}
    \label{fig:all_radar}
\end{figure*}

Next, we provide $MAP@100$ retrieval performance before and after fine-tuning each model (Figure \ref{fig:all_radar}). Table \ref{table:embeddings-results} compares retrieval performance across several industrial domain tasks before and after fine-tuning for a range of embedding models, including BM25, BERT, MPNet, BGE, Qwen2, and E5-Mistral. Metrics reported include HIT@1, MAP@100, and NDCG@10. Across all tasks, fine-tuning significantly boosts performance, especially for HIT@1, where models like E5-Mistral-7B and MPNet-base-v2 often outperform others. Notably, E5-Mistral-7B consistently achieves the highest post-tuning scores across most tasks, indicating strong alignment between embedding quality and domain-specific retrieval needs.

Traditional lexical methods like BM25 perform poorly, especially on HIT@1, highlighting the limitations of non-semantic approaches in complex technical domains. Fine-tuned dense retrievers (e.g., MPNet and BGE) show marked improvements, though smaller models like BERT still lag behind. These results emphasize the importance of model scale, architecture, and task-aware fine-tuning in enhancing semantic retrieval for industrial applications.

\begin{table*}
    \centering
    \resizebox{0.9\textwidth}{!}{%
    \begin{tabular}{|p{1.4cm}|p{5cm}|p{7cm}|p{3cm}|}
    \hline
    \textbf{Task} & \textbf{Example question} & \textbf{Example augmentation} & \textbf{Example answer} \\
    \hline
    A2S & Instruct: What sensors are relevant for the given asset and its category? Query: Asset: aero gas turbine, Category: rotating & Asset description: A rotary machine that extracts energy from steam and converts it into mechanical work & Sensor: amps \\
    \hline
    C2FM & Instruct: Given an asset component, what are the possible failure modes it could experience? Query: Component: Turbo-expanders & Component Description: Turbo-expanders are industrial components that convert the pressure energy of a high-pressure gas into mechanical energy, often used in power generation, refrigeration, and other applications where gas expansion can be harnessed to drive turbines or other machinery. & Failure mode: Failure to set/retrieve (SET)\\
    \hline  
    E2CAT & Instruct: In the context of a given equipment category, what equipment is the most relevant? Query: Equipment category: Electrical & Equipment Category Description: The Electrical equipment category includes a wide range of devices and systems that generate, transmit, distribute, and utilize electrical energy, such as generators, transformers, circuit breakers, and lighting fixtures & Equipment: Coiled tubing, work strings\\
    \hline
    E2CLT & Instruct: For a given equipment class, which types of equipment are most essential? Query: Equipment Class: Swivels & Equipment Class Description: Swivels are industrial components that allow for rotational movement, enabling hoses, pipes, or other equipment to pivot freely while maintaining a secure connection. & Equipment Type: Toxic gases\\
    \hline
    EU2SU & Instruct: What components and their groups are part of a specific equipment unit? Query: Equipment unit: Subsea pipelines & (No augmentation was done$^*$) & Component group: Mounting assembly, Component name: Mounting connector\\
    \hline
    FM2CMP & Instruct: Given the asset name and failure mode class, what components are involved? Query: Asset name: well completion, failure mode class: Low output & (No augmentation was done$^*$) & Component: Top drives\\
    \hline
    FM2CLS & Instruct: For the given failure description, which failure mode class applies? Query: Failure Description: Failed set/retrieve operations & (No augmentation was done$^*$) & Failure class: Power/signal transmission failure\\
        \hline
    FM2S & Instruct: What sensors can be applied to detect a fault in an asset and its category? Query: Asset: electric motor, Category: electric, Fault: stator windings fault & Asset description: Converts electrical energy into mechanical energy to power various industrial machinery.\newline Fault description: A Stator windings fault is a type of industrial failure mode where the electrical windings in the stator of an electric motor or generator become damaged or degraded, often due to overheating, insulation breakdown, or physical stress, leading to reduced performance, efficiency, or complete failure of the equipment. & Sensor: output power \\
    \hline
    S2FM & Instruct: What failure modes can be detected by reading a sensor in an asset and its category? Query: Asset: electric motor, Category: electric, Sensor: current & Asset description: Converts electrical energy into mechanical energy to power various industrial machinery. \newline Sensor Description: Sensor that measures electrical current in various systems to detect anomalies and prevent overloads or system failures & Failure mode: Rotor Windings Fault \\
    \hline
\end{tabular}%
}
    \caption{Example queries, LLM augmentation, and answers for each task. No augmentation was done for queries that we determined that has already enough information to be answered and the performance was already good.}
    \label{tab:task-examples}
\end{table*}

\clearpage
\begin{table*}[ht]
\centering
\begin{adjustbox}{width=0.82\textwidth}
% \resizebox{\textwidth}{!}{%
\begin{tabular}{|c|c|c||l|l||l|l||l|l||}
% \begin{tabular}{ |p{1.7cm}|p{0.7cm}|p{0.7cm}|p{0.7cm}|p{0.7cm}|p{0.7cm}|p{0.7cm}| }
\hline
\multirow{2}{*}{\textbf{Task}} & \multirow{2}{*}{\textbf{Model}} & \multirow{2}{*}{\textbf{Size}} & \multicolumn{2}{c||}{\textbf{HIT@1}} & \multicolumn{2}{c||}{\textbf{MAP@100}} & \multicolumn{2}{c||}{\textbf{NDCG@10}} \\
% \hline
\cline{4-9}
 & & & before & after & before & after & before & after\\
\hline
 \hline
\multirow{6}{2cm}{ Asset to Sensor (A2S)} & BM25 & - & \multicolumn{2}{
c||}{0.00} & \multicolumn{2}{c||}{0.76} & \multicolumn{2}{c||}{2.42}\\
\cline{2-9}
& BERT & 110M & 0.00 & 80.7 & 0.00 & 76.87 & 0.00 & 80.66\\
\cline{2-9}
& MPNet-base-v2 & 110M & 5.26 & \textbf{91.23} & 1.69 & 80.94 & 2.47 & 85.19\\
\cline{2-9}
& BGE-Large-v1.5 & 335M & 0.00 & 84.21 & 5.20 & 63.81 & 2.77 & 69.05\\  
\cline{2-9}
 &  Qwen2-7B-Instruct & 7B & 0.00 & 84.21 & 0.16 & 73.40 & 0.00 & 78.27\\ 
\cline{2-9}
&  E5-Mistral-7B & 7B & \textbf{15.79} & \textbf{91.23} & \textbf{14.43} & \textbf{84.62} & \textbf{18.46} & \textbf{87.03}\\
 \hline
 \hline
\multirow{6}{2cm}{Component to Failure Mode (C2FM)} & BM25 & - & \multicolumn{2}{
c||}{0.00} & \multicolumn{2}{c||}{0.30} & \multicolumn{2}{c||}{0.73}\\
\cline{2-9}
\cline{2-9}
 & BERT & 110M & 0.00 & 4.22 & 0.00 & 10.89 & 0.00 & 13.46\\
\cline{2-9}
& MPNet-base-v2 & 110M & 0.00 & 9.64 & 1.48 & 17.6 & \textbf{1.87} & 21.54\\
\cline{2-9}
& BGE-Large-v1.5 & 335M & 0.00 & 3.61 & 1.71 & 10.91 & 0.95 & 13.96\\  
\cline{2-9}
 &  Qwen2-7B-Instruct & 7B & 0.00 & 8.43 & 0.02 & 14.33 & 0.00 & 18.79\\ 
\cline{2-9}
&  E5-Mistral-7B & 7B & 0.00 & \textbf{9.64} & \textbf{2.18} & \textbf{26.20} & 1.83 & \textbf{32.50}\\
 \hline
 \hline
\multirow{6}{2cm}{Equipment to Category (E2CAT)} & BM25 & - & \multicolumn{2}{
c||}{14.29} & \multicolumn{2}{c||}{4.61} & \multicolumn{2}{c||}{12.53}\\
\cline{2-9}
& BERT & 110M & 0.00 & 77.78 & 0.00 & 76.17 & 0.00 & 86.78\\
\cline{2-9}
& MPNet-base-v2 & 110M & 0.00 & \textbf{100.0} & 2.82 & 81.83 & 6.99 & 88.95\\
\cline{2-9}
& BGE-Large-v1.5 & 335M & 22.22 & 88.89 & 4.59 & \textbf{83.46} & 9.80 & \textbf{90.90}\\  
\cline{2-9}
 &  Qwen2-7B-Instruct & 7B & 0.00 & 88.89 & 5.37 & 68.82 & 3.56 & 74.38\\ 
\cline{2-9}
&  E5-Mistral-7B & 7B & \textbf{44.44} & 77.78 & \textbf{41.87} & 70.80 & \textbf{54.70} & 83.61\\
\hline
\hline
\multirow{6}{2cm}{Equipment to Class Type (E2CLT)} & BM25 & - & \multicolumn{2}{
c||}{0.00} & \multicolumn{2}{c||}{1.08} & \multicolumn{2}{c||}{00.00}\\
\cline{2-9}
& BERT & 110M & 0.00 & \textbf{85.71} & 0.00 & 82.22 & 0.00 & 83.30\\
\cline{2-9}
& MPNet-base-v2 & 110M & 2.38 & 73.81 & \textbf{15.48} & 75.07 & \textbf{18.83} & 78.39\\
\cline{2-9}
& BGE-Large-v1.5 & 335M & \textbf{7.14} & \textbf{85.71} & 14.12 & \textbf{82.89} & 16.2 & \textbf{83.99}\\  
\cline{2-9}
 &  Qwen2-7B-Instruct & 7B & 0.00 & 66.67 & 6.33 & 61.19 & 9.25 & 64.54\\ 
\cline{2-9}
&  E5-Mistral-7B & 7B & 0.00 & 73.81 & 12.76 & 70.41 & 16.43 & 69.61\\
 \hline
\hline
\multirow{6}{2cm}{Equipment Unit to Subunit (EU2SU)} & BM25 & - & \multicolumn{2}{
c||}{0.00} & \multicolumn{2}{c||}{0.53} & \multicolumn{2}{c||}{0.55}\\
\cline{2-9}
& BERT & 110M & 0.00 & \textbf{100.0} & 1.35 & \textbf{91.56} & 1.53 & \textbf{95.91}\\
\cline{2-9}
& MPNet-base-v2 & 110M & \textbf{14.63} & \textbf{100.0} & 7.34 & 88.43 & 14.95 & 92.84\\
\cline{2-9}
& BGE-Large-v1.5 & 335M & 7.32 & 92.68 & 8.03 & 90.79 & \textbf{16.24} & 91.43\\  
\cline{2-9}
 &  Qwen2-7B-Instruct & 7B & 0.00 & 95.12 & 1.52 & 88.06 & 4.15 & 93.94\\ 
\cline{2-9}
&  E5-Mistral-7B & 7B & 0.00 & 92.68 & \textbf{8.88} & 84.63 & 13.85 & 88.74\\
 \hline
\hline
\multirow{6}{2cm}{Failure Mode to Components (FM2CMP)} & BM25 & - & \multicolumn{2}{
c||}{0.00} & \multicolumn{2}{c||}{0.00} & \multicolumn{2}{c||}{0.00}\\
\cline{2-9}
& BERT & 110M & 0.00 & 48.42 & 0.00 & 60.89 & 0.00 & 69.04\\
\cline{2-9}
& MPNet-base-v2 & 110M & 0.00 & 61.99 & 0.86 & 64.43 & 0.93 & 70.40\\
\cline{2-9}
& BGE-Large-v1.5 & 335M & 0.00 & 54.75 & 0.23 & 63.28 & 0.00 & 69.82\\  
\cline{2-9}
 &  Qwen2-7B-Instruct & 7B & 0.00 & \textbf{70.59} & 0.57 & \textbf{74.86} & 0.94 & \textbf{78.51}\\ 
\cline{2-9}
&  E5-Mistral-7B & 7B & \textbf{4.07} & 59.73 & \textbf{10.94} & 69.62 & \textbf{13.69} & 74.26\\
 \hline
 \hline
\multirow{6}{2cm}{Failure Mode to Class (FM2CLS)} & BM25 & - & \multicolumn{2}{
c||}{8.14} & \multicolumn{2}{c||}{12.78} & \multicolumn{2}{c||}{14.55}\\
\cline{2-9}
& BERT & 110M & 3.23 & 82.26 & 4.62 & 84.41 & 4.57 & 84.97\\
\cline{2-9}
& MPNet-base-v2 & 110M & 20.43 & \textbf{82.8} & 29.31 & \textbf{84.68} & 31.94 & \textbf{85.17}\\
\cline{2-9}
& BGE-Large-v1.5 & 335M & \textbf{34.41} & 81.72 & \textbf{45.07} & 84.20 & \textbf{50.04} & 84.85\\  
\cline{2-9}
 &  Qwen2-7B-Instruct & 7B & 2.15 & 77.96 & 9.96 & 82.13 & 12.69 & 83.18\\ 
\cline{2-9}
&  E5-Mistral-7B & 7B & 5.91 & \textbf{82.8} & 30.08 & 83.14 & 39.3 & 83.98\\
 \hline
 \hline
\multirow{6}{2cm}{Failure Mode to Sensor (FM2S)} & BM25 & - & \multicolumn{2}{
c||}{0.00} & \multicolumn{2}{c||}{2.84} & \multicolumn{2}{c||}{4.45}\\
\cline{2-9}
& BERT & 110M & 0.00 & 38.38 & 0.00 & 41.71 & 0.00 & 50.70\\
\cline{2-9}
& MPNet-base-v2 & 110M & 0.00 & \textbf{49.49} & 0.70 & 47.81 & 0.63 & 59.48\\
\cline{2-9}
& BGE-Large-v1.5 & 335M & 0.00 & 48.48 & 2.77 & 40.03 & 2.77 & 48.45\\  
\cline{2-9}
 &  Qwen2-7B-Instruct & 7B & 0.00 & 42.42 & 2.10 & 39.39 & 4.10 & 46.87\\ 
\cline{2-9}
&  E5-Mistral-7B & 7B & \textbf{3.03} & 48.48 & \textbf{5.38} & \textbf{54.27} & \textbf{7.73} & \textbf{61.69}\\
 \hline
 \hline
\multirow{6}{2cm}{Sensor to Failure Modes (S2FM)} & BM25 & - & \multicolumn{2}{
c||}{00.00} & \multicolumn{2}{c||}{0.46} & \multicolumn{2}{c||}{0.00}\\
\cline{2-9}
& BERT & 110M & 0.00 & 11.48 & 0.00 & 20.76 & 0.00 & 25.04\\
\cline{2-9}
& MPNet-base-v2 & 110M & 0.00 & 19.34 & 0.03 & 36.76 & 0.00 & \textbf{44.36}\\
\cline{2-9}
& BGE-Large-v1.5 & 335M & 0.00 & 27.54 & 0.66 & 38.12 & 0.85 & 42.52\\  
\cline{2-9}
 &  Qwen2-7B-Instruct & 7B & 0.00 & 20.33 & 0.36 & 32.41 & 0.27 & 37.32\\ 
\cline{2-9}
&  E5-Mistral-7B & 7B & 0.00 & \textbf{33.77} & \textbf{1.57} & \textbf{39.07} & \textbf{1.55} & 43.81\\
 \hline
\end{tabular}
% }
\end{adjustbox}
    \caption{Retrieval performance before and after fine-tuning. For tasks with a high average number of related items per query, the HIT score is naturally higher.}
    \label{table:embeddings-results}
\end{table*}
\clearpage
% \section{Training Domain-Specific Embedder from Web Data for a more Robust Baseline}
\section{Towards a more Robust Baseline Domain Embedder from Web Data}
The pre-trained embedders didn't show a satisfactory performance (Table \ref{table:embeddings-results}). In our effort to build a more robust baseline, we use an agent-driven domain web data collection approach and fine-tune using Masked Language Modeling (MLM).

\subsection{Dataset Preparation}
We set up a ReAct agent \cite{yao2022react} equipped with two tools that connect to external knowledge sources: ArXiv \cite{langchain_arxiv_tool} and Wikipedia \cite{langchain_wikipedia_tool}. We then use a series of multiple-choice industrial questions from the \texttt{FailureSensorIQ} dataset \cite{constantinides2025failuresensoriq} and employ the ReAct agent to generate answers by leveraging information retrieved from these external knowledge bases. During execution, all interactions between the ReAct agent and the external tools are stored as key-value pairs. 
% This key-value store enables us to systematically trace the agent’s reasoning process, facilitate answer verification, and support subsequent analysis or training data generation. 
Table \ref{tab:rotor_windings_fault_relevance} provides an example of the key-value store for Wikipedia, focusing on a particular key named ``Rotor windings fault in electric motors''. We treat the values as unique passages for fine-tuning an embedding model tailored to our domain. Overall, we collected 10552 passages from Wikipedia and 11515 passages from ArXiv.

\subsection{Model Training and Experimental Results}
We train different models using Masked Language Modeling (MLM) on the collected passages and report the results in Table \ref{tab:mlm}. We train for 100 epochs on a machine with 1 Nvidia A100 (80GB), using a batch size of 32 and learning rate of $2*10^{-5}$, weight decay of $0.01$, and token masking probability of $15\%$. Overall, the performance is still unsatisfactory. The key takeaways are: (a) carefully curated datasets with domain-specific instructions based on day-to-day operations are crucial for learning good embeddings, and (b) publicly available web documents lack sufficient details on the relationships necessary to model interactions between different industrial entities (e.g., sensors, failure modes, components).

\begin{table}[h]
\centering
\begin{tabular}{|p{2.5cm}|p{4.2cm}|}
\hline
\textbf{Motor Type (Wikipedia)} & \textbf{Rotor Windings Fault Relevance Summary} \\ \hline
Reluctance Motor (Reluctance motor) & Rotor does \textbf{not have windings}; torque is generated via magnetic reluctance. \textbf{Not relevant} to rotor winding faults. \\ \hline
Brushed DC Motor (Brushed DC electric motor) & Rotor includes windings and uses brushes for commutation. Wear and tear may impact windings. \textbf{Relevant} to rotor winding fault scenarios. \\ \hline
Doubly Fed Induction Generator (Doubly fed electric machine) & Rotor has \textbf{field windings} connected to external circuits; faults in rotor windings can affect performance. \textbf{Highly relevant} to rotor winding fault analysis. \\ \hline
\end{tabular}
\caption{Summary of rotor windings fault relevance for different electric motor types from Wikipedia pages}
\label{tab:rotor_windings_fault_relevance}
\end{table}

% lack of purpose build corpus

% future work - its does not exists explicit, it can be treated as grounded document for the type of task we want to do. 

% in general these embedder are missing domain specific instructions used in day 2 day operation. 

\begin{table*}
\centering
\setlength{\tabcolsep}{8pt} % Column separation
\renewcommand{\arraystretch}{1.2} % Row height
\begin{tabular}{@{} 
    p{3.5cm} c || cc || cc || cc @{}}
\toprule
\multirow{2}{*}{\textbf{Task}} & \multirow{2}{*}{\textbf{Model}} 
& \multicolumn{2}{c||}{\textbf{ACC@1}} 
& \multicolumn{2}{c||}{\textbf{MAP@100}} 
& \multicolumn{2}{c@{}}{\textbf{NDCG@10}} \\
\cmidrule(r){3-4} \cmidrule(r){5-6} \cmidrule(l){7-8}
 & & Before & After & Before & After & Before & After \\
\midrule
\multirow{3}{3.5cm}{\textbf{Asset to Sensor (A2S)}} 
& BERT & 0.00 & 0.00 & 0.00 & 0.00 & 0.00 & 0.00 \\
& MPNET & 0.00 & 0.00 & \textbf{1.51} & 0.00 & \textbf{0.71} & 0.00 \\
& BGE & 0.00 & 0.00 & \textbf{6.36} & 0.00 & 0.00 & 0.00 \\
\midrule
\multirow{3}{3.5cm}{\textbf{Component to Failure Mode (C2FM)}} 
& BERT & 0.00 & 0.00 & 0.00 & \textbf{0.16} & 0.00 & \textbf{1.49} \\
& MPNET & 0.00 & 0.00 & 0.00 & 0.00 & 0.00 & 0.00 \\
& BGE & 0.00 & 0.00 & \textbf{0.31} & 0.00 & 0.00 & 0.00 \\
\midrule
\multirow{3}{3.5cm}{\textbf{Equipment to Category (E2CAT)}} 
& BERT & 0.00 & \textbf{16.67} & 1.38 & \textbf{3.18} & 3.52 & \textbf{9.32} \\
& MPNET & 0.00 & 0.00 & 0.41 & \textbf{2.13} & 5.77 & \textbf{6.22} \\
& BGE & \textbf{16.7} & 0.00 & \textbf{6.80} & 0.75 & \textbf{12.2} & 0.00 \\
\midrule
\multirow{3}{3.5cm}{\textbf{Equipment to Class Type (E2CLT)}} 
& BERT & \textbf{4.88} & 0.00 & 1.74 & \textbf{1.99} & \textbf{3.67} & 1.14 \\
& MPNET & 0.00 & 0.00 & 7.13 & \textbf{12.5} & \textbf{11.80} & 4.58 \\
& BGE & 2.44 & \textbf{19.5} & 4.88 & \textbf{6.86} & 4.99 & \textbf{9.43} \\
\midrule
\multirow{3}{3.5cm}{\textbf{Equipment Unit to Subunit (EU2SU)}} 
& BERT & 2.63 & \textbf{10.53} & 1.97 & \textbf{4.96} & 5.67 & \textbf{13.57} \\
& MPNET & \textbf{15.79} & 0.00 & \textbf{8.92} & 4.58 & \textbf{22.68} & 12.47 \\
& BGE & 18.4 & \textbf{31.58} & 10.96 & \textbf{11.94} & 24.11 & \textbf{31.18} \\
\midrule
\multirow{3}{3.5cm}{\textbf{Failure Mode to Components (FM2CMP)}} 
& BERT & 0.00 & 0.00 & \textbf{0.28} & 0.00 & \textbf{0.31} & 0.00 \\
& MPNET & \textbf{0.45} & 0.00 & \textbf{2.70} & 0.27 & \textbf{3.34} & 0.53 \\
& BGE & 0.00 & 0.00 & \textbf{0.92} & 0.00 & \textbf{10.57} & 0.00 \\
\midrule
\multirow{3}{3.5cm}{\textbf{Failure Mode to Class (FM2CLS)}} 
& BERT & 0.00 & \textbf{5.08} & 1.15 & \textbf{12.7} & 0.60 & \textbf{16.5} \\
& MPNET & \textbf{20.34} & 20.34 & 30.80 & \textbf{33.37} & 34.68 & \textbf{37.52} \\
& BGE & \textbf{29.66} & 28.81 & \textbf{37.85} & 35.85 & \textbf{41.31} & 38.22 \\
\midrule
\multirow{3}{3.5cm}{\textbf{Failure Mode to Sensor (FM2S)}} 
& BERT & 0.00 & 0.00 & 0.00 & \textbf{2.1} & 0.00 & \textbf{0.8} \\
& MPNET & 0.00 & 0.00 & 0.23 & \textbf{0.83} & 0.00 & \textbf{0.13} \\
& BGE & 0.00 & 0.00 & \textbf{4.25} & 0.00 & \textbf{2.27} & 0.00 \\
\midrule
\multirow{3}{3.5cm}{\textbf{Sensor to Failure Modes (S2FM)}} 
& BERT & 0.00 & \textbf{2.13} & 0.82 & \textbf{3.46} & 0.78 & \textbf{3.31} \\
& MPNET & \textbf{2.13} & 0.00 & \textbf{0.36} & 0.00 & \textbf{0.66} & 0.00 \\
& BGE & \textbf{0.71} & 0.00 & \textbf{3.65} & 0.00 & \textbf{4.05} & 0.00 \\
\bottomrule
\end{tabular}
\caption{Performance before and after Masked Language Modeling (MLM) on industrial web documents. Overall, performance remains poor and improvements are inconsistent. This underscores the importance of carefully curating instruction-based datasets tailored to specific industrial tasks, as demonstrated in our methodology.}
\label{tab:mlm}
\end{table*}
% \clearpage
\section{Hallucination Analysis on the LLM Augmentation}
As discussed in Section~\ref{llm-query-aug}, our approach leverages an LLM to augment queries, making it crucial to assess the quality of the generated text by LLM.

\begin{figure}[h!]
    \centering
    \includegraphics[width=1\linewidth]{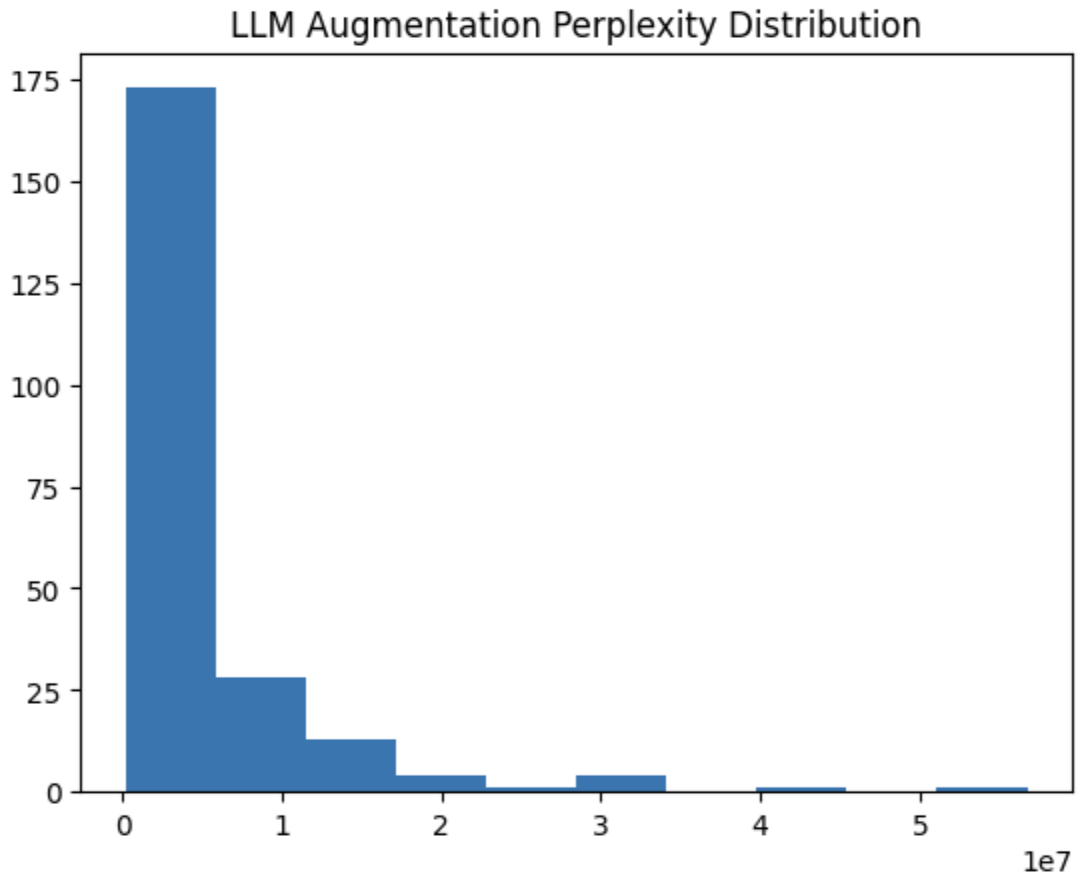}
    \caption{Perplexity Distribution of the augmented entity descriptions using Llama-3.3-70B-Instruct.}
    \label{fig:ppl}
\end{figure}

\begin{figure}[h!]
    \centering
    \includegraphics[width=1\linewidth]{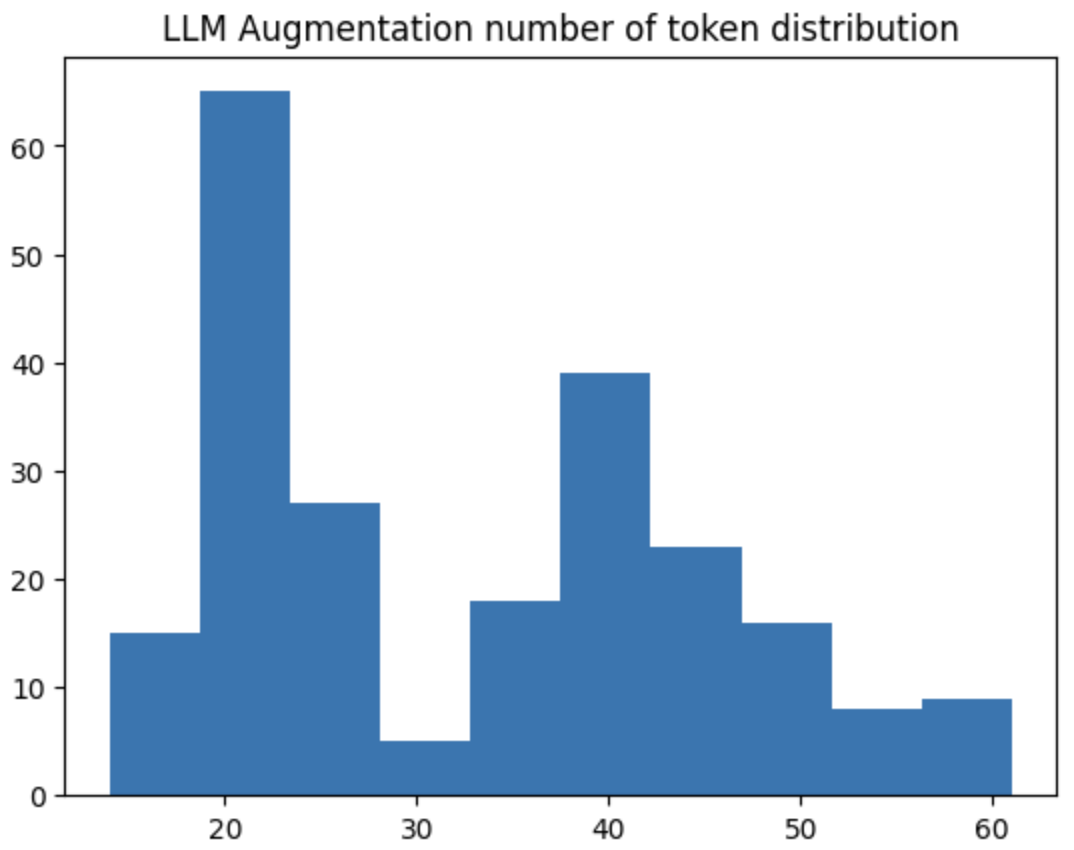}
    \caption{Number of Tokens Distribution with the augmented entity descriptions using Llama-3.3-70B-Instruct.}
    \label{fig:tokens}
\end{figure}

We extract all the LLM generated text to conduct deeper study. All together, the dataset includes 255 unique entities spanning assets, sensors, failure modes, equipment components, categories, class types, subunits, and units and we have 225 summary in a form of single sentence is extracted. We analyze the distribution of token lengths and perplexity scores for entity descriptions generated using LLaMA-3.3-70B-Instruct (Figures~\ref{fig:ppl} and~\ref{fig:tokens}). The token counts follow an approximately bimodal normal distribution, while perplexity exhibits a heavy right-tailed distribution. 
% Some sentences about how did we calculated perplexity? if any one like to produce.

To further evaluate quality, we manually inspect generated descriptions in the top 5\% percentile of perplexity scores for factual accuracy (Example descriptions in Table \ref{tab:task-examples} as an example). Although these prompts are found to be factually correct, token count and perplexity may still serve as useful signals for identifying potential hallucinations and warrant further investigation.

\begin{table}[ht!]
\centering
\label{tab:groundedness_stats}
\begin{tabular}{@{}lr@{}}
\toprule
\textbf{Metric} & \textbf{Value} \\
\midrule
Total Evaluations & 222 \\
Mean Score       & 0.874 \\
Median Score     & 0.800 \\
Standard Deviation & 0.101 \\
Minimum Score    & 0.5 \\
Maximum Score    & 1.0 \\
\midrule
\multicolumn{2}{@{}l}{\textbf{Distribution (Score Range \(\to\) Count)}} \\
\midrule
0.0 -- 0.2 & 0 \\
0.2 -- 0.4 & 0 \\
0.4 -- 0.6 & 2 \\
0.6 -- 0.8 & 5 \\
0.8 -- 1.0 & 215 \\
\bottomrule
\end{tabular}
\caption{Summary Statistics and Distribution of Groundedness Scores}
\label{rscore}
\end{table}

\subsection{FactChecker Agent}

To validate the factual consistency of textual summary, we develop a ReAct-based fact-checking agent that performs retrieval-augmented verification. For each summary, the agent queries authoritative sources such as Wikipedia and arXiv to gather evidence that either supports or contradicts the claim. As specified in Table~\ref{tab:factcheckerprompt}, the protocol enables fine-grained evaluation by aggregating evidence across steps, assigning confidence scores, and detecting contradictions. This dual-source validation strategy culminates in a groundedness score for the overall rationale, offering a robust assessment of factuality in generated explanations.

We evaluate the factual grounding of 222 agent-generated summary using a groundedness scoring metric ranging from 0.0 (no support) to 1.0 (strong support). Table \ref{rscore} show the outcome of our experiment. The analysis reveals a high overall factual accuracy, with a mean score of 0.874 and a median of 0.800, indicating that most rationales are well-supported by reliable technical sources. The standard deviation of 0.101 reflects moderate variability, while the minimum and maximum scores were 0.5 and 1.0, respectively. Distribution analysis shows that the vast majority (215 out of 222) of scores fall within the 0.8 to 1.0 range, confirming strong evidence backing the agents’ conclusions. Only a small fraction exhibits weaker grounding, highlighting areas for potential improvement.

\begin{figure*}[ht!]
\centering
\begin{tabular}{|p{15cm}|}
\hline
{\ttfamily
Task: Assess whether the following step-by-step rationale about an industrial asset and its associated sensor-based failure mode identification is \textbf{factually grounded} in reliable technical or scientific sources.\newline\newline
Your role is that of a Reliability Engineering Expert specializing in sensor-based condition monitoring and failure diagnostics across a wide range of industrial assets (e.g., turbines, pumps, motors, HVAC systems, rotating machinery, etc.).\newline\newline
Context:\newline
- Asset Type: "\{asset\_type\}"\newline
- Sensor Type: "\{sensor\_type\}"\newline
- Rationale: "\{rationale\_text\}"\newline\newline
Instructions:\newline
1. Search reliable sources such as \textbf{Wikipedia}, \textbf{arXiv}, and other authoritative engineering or maintenance references to find passages that either \textbf{support} or \textbf{contradict} the claims made in each step of the rationale.\newline
2. For each reasoning step, assess whether the technical claim is:\newline
\hspace{0.5cm} - \texttt{correct}: factually grounded and logically sound.\newline
\hspace{0.5cm} - \texttt{partially\_correct}: partially grounded but includes gaps or weak assumptions.\newline
\hspace{0.5cm} - \texttt{incorrect}: not supported or contradicted by reliable sources.\newline
3. For each step:\newline
\hspace{0.5cm} - Provide a \textbf{confidence score} (0.0–1.0) reflecting your certainty.\newline
\hspace{0.5cm} - Provide a \textbf{brief comment} justifying your assessment.\newline
\hspace{0.5cm} - Include \textbf{any relevant supporting or contradicting passages} you find from external sources with \textbf{citations}.\newline
...
5. Assign a \textbf{groundedness score} to the entire rationale, from \texttt{0.0} (no support or contradicted) to \texttt{1.0} (strongly supported).\newline
6. You must conclude with a \texttt{Finish} action that returns a fully filled, valid, and parseable \textbf{JSON object} matching the exact structure below.\newline
\hspace{0.5cm} - Do \textbf{not} use placeholders.\newline
\hspace{0.5cm} - The process is \textbf{not complete} unless a proper JSON is returned.\newline\newline
Output Format (JSON):\newline
\{\newline
\hspace{0.5cm} "asset\_type": "\{asset\_type\}",\newline
\hspace{0.5cm} "sensor\_type": "\{sensor\_type\}",\newline
\hspace{0.5cm} "rationale": "\{rationale\_text\}",\newline
\hspace{0.5cm} "evaluation": [\newline
\hspace{1cm} \{\newline
\hspace{1.5cm} "step": 1,\newline
\hspace{1.5cm} "status": "correct" | "partially\_correct" | "incorrect",\newline
\hspace{1.5cm} "comment": "...",\newline
\hspace{1.5cm} ],\newline
\hspace{1.5cm} "contradicting\_passages": [\newline
.....suppressed....
\hspace{1cm} "justification": "...",\newline
\hspace{1cm} "groundedness\_score": 0.0\newline
\hspace{0.5cm} \}\newline
\}\newline
(END OF RESPONSE)
}\\
\hline
\end{tabular}
\caption{Sensor-Groundedness Evaluation Prompt}
\label{tab:factcheckerprompt}
\end{figure*}

% \resizebox{.4\textwidth}{!}{
%     \begin{tcolorbox}[colframe=black, colback=white, sharp corners, boxrule=0.8mm]
%         \begin{tabular}{p{0.45\textwidth} p{0.45\textwidth}}
%             \rowcolor{lightgreen!30!white} \textbf{Instruct:} What sensors can be applied to detect a fault in an asset and its category? & \textbf{Query:} \underline{Asset}: electric motor, \underline{Category}: electric, \underline{Fault:} stator windings fault \\
%             \rowcolor{customblue!10!white} \textbf{Asset description:} Converts electrical energy into mechanical energy to power various industrial machinery. & \textbf{Fault description:} A Stator windings fault is a type of industrial failure mode where the electrical windings in the stator of an electric motor or generator become damaged or degraded, often due to overheating, insulation breakdown, or physical stress, leading to reduced performance, efficiency, or complete failure of the equipment. \\
%             \rowcolor{pink!10!white} \multicolumn{2}{l}{\textbf{Sensor:} output power}
%         \end{tabular}
%     \end{tcolorbox}
% }
\clearpage
 \end{document}